\documentclass[runningheads]{llncs}

\usepackage[year=2026]{eccv}
\usepackage{eccvabbrv}
\usepackage{graphicx}
\usepackage{booktabs}
\usepackage{subcaption}
\usepackage{multirow}
\usepackage{amsmath}

\usepackage{xcolor}
\usepackage{tikz}
\usetikzlibrary{arrows.meta,positioning,calc,fit,backgrounds}

\usepackage{listings}
\lstdefinestyle{prompt}{
  basicstyle=\ttfamily\footnotesize,
  breaklines=true, breakindent=0pt,
  columns=fullflexible, keepspaces=true,
  frame=single, framesep=4pt, xleftmargin=2pt,
}

\usepackage{newunicodechar}
\newunicodechar{°}{\ensuremath{^\circ}}\newunicodechar{×}{\ensuremath{\times}}\newunicodechar{–}{--}\newunicodechar{—}{---}
\newunicodechar{↔}{\ensuremath{\leftrightarrow}}\newunicodechar{δ}{\ensuremath{\delta}}\newunicodechar{α}{\ensuremath{\alpha}}
\newunicodechar{≈}{\ensuremath{\approx}}\newunicodechar{→}{\ensuremath{\rightarrow}}\newunicodechar{≤}{\ensuremath{\le}}
\newunicodechar{≥}{\ensuremath{\ge}}\newunicodechar{≪}{\ensuremath{\ll}}\newunicodechar{≫}{\ensuremath{\gg}}\newunicodechar{·}{\ensuremath{\cdot}}
\newunicodechar{∈}{\ensuremath{\in}}\newunicodechar{±}{\ensuremath{\pm}}\newunicodechar{−}{\ensuremath{-}}\newunicodechar{≠}{\ensuremath{\neq}}
\newunicodechar{“}{``}\newunicodechar{”}{''}\newunicodechar{‘}{`}\newunicodechar{’}{'}\newunicodechar{Δ}{\ensuremath{\Delta}}\newunicodechar{★}{\ensuremath{\star}}\newunicodechar{∼}{\ensuremath{\sim}}\newunicodechar{Φ}{\ensuremath{\Phi}}\newunicodechar{σ}{\ensuremath{\sigma}}\newunicodechar{ρ}{\ensuremath{\rho}}\newunicodechar{θ}{\ensuremath{\theta}}\newunicodechar{ℓ}{\ensuremath{\ell}}
\newunicodechar{é}{\'e}\newunicodechar{è}{\`e}\newunicodechar{ê}{\^e}\newunicodechar{ô}{\^o}\newunicodechar{ö}{\"o}\newunicodechar{ü}{\"u}\newunicodechar{ä}{\"a}\newunicodechar{á}{\'a}\newunicodechar{ñ}{\~n}\newunicodechar{ç}{\c{c}}

\usepackage{hyperref}

\definecolor{loopInk}{HTML}{1F2933}
\definecolor{loopSlate}{HTML}{5B6B7B}
\definecolor{loopSlateBg}{HTML}{F0F3F6}
\definecolor{loopBlue}{HTML}{2C5F8A}
\definecolor{loopBlueBg}{HTML}{E8F0F7}
\definecolor{loopAmber}{HTML}{9C6314}
\definecolor{loopAmberBg}{HTML}{FBF1E0}
\definecolor{loopGreen}{HTML}{2E6B4F}
\definecolor{loopGreenBg}{HTML}{E8F2EC}
\definecolor{loopRed}{HTML}{A33B3B}
\definecolor{loopRedBg}{HTML}{FAECEC}

\newcommand{\HumanTFPone}{0.49}
\newcommand{\HumanTFPend}{0.93}
\newcommand{\HumanNumFixTP}{3}
\newcommand{\HumanNumFixTA}{5}

\newcommand{\HumanCeiling}{0.53}
\newcommand{\HumanDprime}{2.91}
\newcommand{\NImages}{285}
\newcommand{\NImagesTP}{141}
\newcommand{\NImagesTA}{144}
\newcommand{\NSeeds}{5}
\newcommand{\PCApcone}{45}
\newcommand{\PCApctwo}{30}
\newcommand{\NEpisodes}{38{,}475}
\newcommand{\QwenTPexist}{0.99}
\newcommand{\QwenTAexist}{0.81}
\newcommand{\QwenYesBias}{0.19}

\newcommand{\QwenSharpTFPone}{0.97}

\newcommand{\QwenGPTFPone}{0.97}
\newcommand{\QwenGPTFPend}{0.98}
\newcommand{\QwenGPNumFixTP}{2}
\newcommand{\QwenGPNumFixTA}{1}
\newcommand{\QwenGPDeclAbs}{0.93}

\newcommand{\QwenGPDprime}{3.84}
\newcommand{\QwenGPEntropyD}{$-$.67}
\newcommand{\QwenGPSaccD}{$+$.50}

\newcommand{\QwenGPSelfConsist}{0.84}

\newcommand{\QwenGPDensityCC}{0.58}

\newcommand{\QwenKthirtytwoTFPone}{0.42}
\newcommand{\QwenKthirtytwoTFPend}{0.71}

\newcommand{\GlmTPexist}{0.99}
\newcommand{\GlmTAexist}{0.81}
\newcommand{\GlmYesBias}{0.19}

\newcommand{\GlmGPTFPone}{0.97}
\newcommand{\GlmGPTFPend}{0.98}
\newcommand{\GlmGPNumFixTP}{2}
\newcommand{\GlmGPNumFixTA}{4}
\newcommand{\GlmGPDeclAbs}{0.91}

\newcommand{\GlmGPDprime}{4.23}
\newcommand{\GlmGPEntropyD}{$-$.65}
\newcommand{\GlmGPSaccD}{$+$.61}

\newcommand{\GlmGPSelfConsist}{0.91}

\newcommand{\GlmGPDensityCC}{0.63}

\newcommand{\GemmaTPexist}{0.97}
\newcommand{\GemmaTAexist}{0.80}
\newcommand{\GemmaYesBias}{0.20}

\newcommand{\GemmaGPTFPone}{0.80}
\newcommand{\GemmaGPTFPend}{0.93}
\newcommand{\GemmaGPNumFixTP}{3}
\newcommand{\GemmaGPNumFixTA}{5}
\newcommand{\GemmaGPDeclAbs}{0.96}

\newcommand{\GemmaGPDprime}{3.14}
\newcommand{\GemmaGPEntropyD}{$-$.27}
\newcommand{\GemmaGPSaccD}{$+$.23}

\newcommand{\GemmaGPSelfConsist}{0.71}

\newcommand{\GemmaGPDensityCC}{0.50}
\newcommand{\GemmaKsixteenTFPone}{0.42}
\newcommand{\GemmaKsixteenTFPend}{0.75}

\newcommand{\GemmaNumFixDelta}{-0.03}

\begin{document}

\title{Matched Outcomes, Divergent Gaze: How Foveated MLLMs Search Compared to Humans \thanks{The recorded data and the analysis code required to reproduce every result will be released upon publication.}}

\title{Matched Outcomes, Divergent Gaze: How Foveated MLLMs Search Compared to Humans \thanks{Code available at: \href{https://github.com/kmamine/MODG}{https://github.com/kmamine/MODG} .}}

\titlerunning{Matched Outcomes, Divergent Gaze}

\author{Mohamed Amine KERKOURI \inst{1} \and Marouane TLIBA\inst{2} \and Aladine CHETOUANI\inst{2} \and Ulas BAGCI \inst{3} \and Alessandro BRUNO \inst{4} }

\authorrunning{Kerkouri et al.}

\institute{F-initiatives, Paris, France\\ \and
  Université Sorbonne Paris Nord, villetaneuse, France\\
  \and Northwestern University, Chicago , IL, USA \\ 
  \and IULM university, Milan, Italy
}

\maketitle

\begin{abstract}
Human visual search is \emph{serial}: the fovea must land on a candidate to confirm it, and those landings form a scanpath. Whether multimodal large language models (MLLMs), given the same foveated input, search as humans do bears on their use as models of human vision and on attention-alignment scores. We compare three general-purpose MLLMs with human eye-movement scanpaths on goal-directed search (COCO-Search18), driving each model fixation by fixation through an identical, human-matched foveated view and assessing it along three axes: the \emph{decision} of target presence, the \emph{efficiency} of reaching the target, and the \emph{gaze process} itself.
The axes dissociate. On the decision and on target acquisition the models match or exceed humans, detecting present targets near ceiling and reaching them on the first saccade more often than people do. The gaze process is not human. Under the human-matched condition, all three share one signature: low-entropy, large-amplitude, self-consistent scanpaths that agree with themselves far more closely than two humans agree with each other. That is consistent with a single-pass, non-serial architecture rather than a limit of acuity. Matched retinal input reproduces where humans look but not how the looking unfolds in time, and no degradation regime recovers human-like search at human-like success. The gap sits on a process axis that answer-alignment and saliency metrics do not measure. Because they miss it, such metrics cannot certify human-like vision, and zero-shot models suit outcome and spatial questions but not temporal, process-level ones.

\end{abstract}

\section{Introduction}

\begin{figure}\centering
  \includegraphics[width=\linewidth,  trim={0 0 0 120pt}, clip]{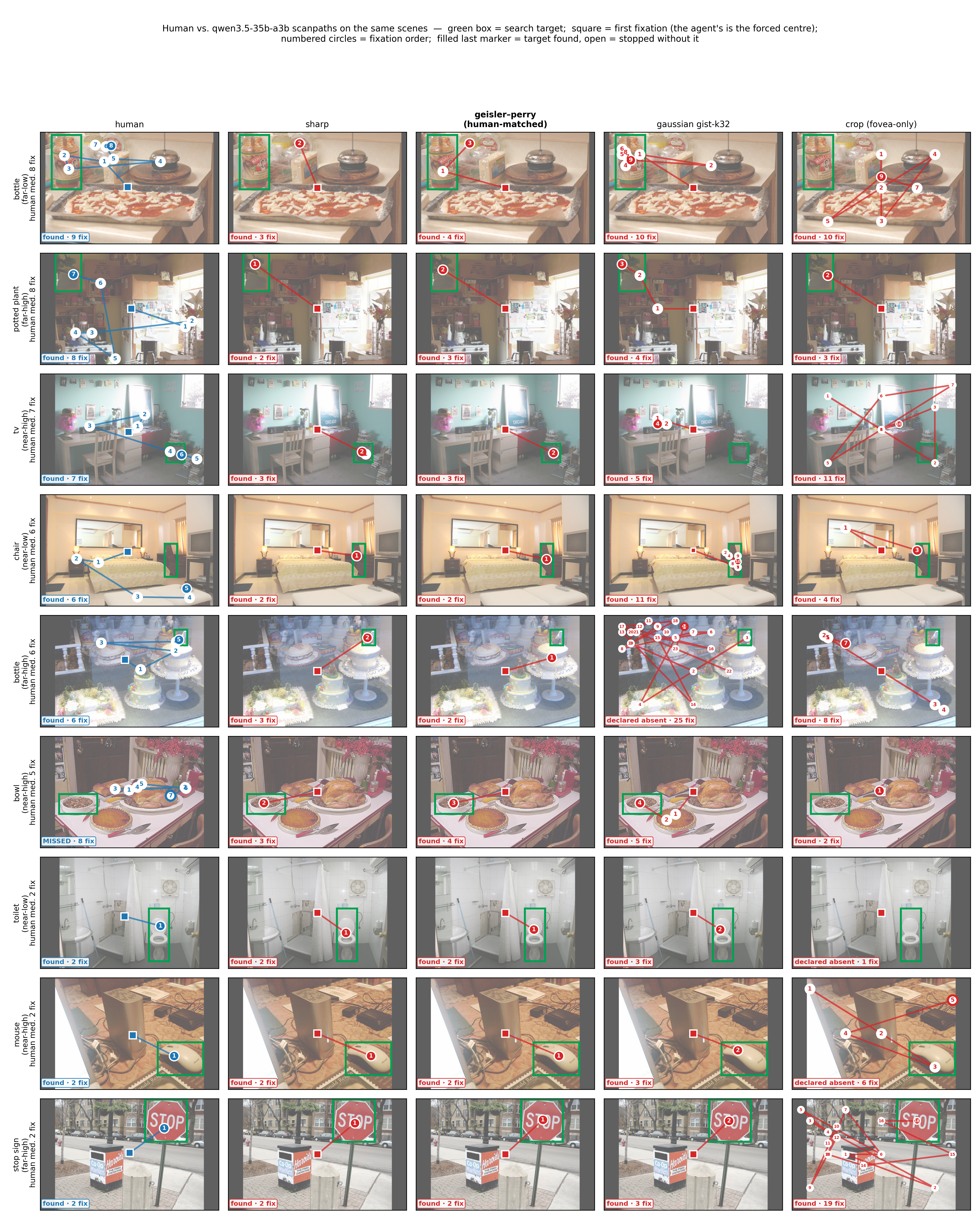}
  \caption{The dissociation on individual trials. Rows are (scene, target) target-present trials ordered by human search difficulty; columns are one human  observer and Qwen3.5-35B-A3B (seed~$0$) under four conditions. Green box, target; square, first fixation; numbered circles, fixation order; filled final marker, target found. Where the human accumulates fixations across the scene before
  confirming the target, the model under \textsc{sharp} and \textsc{geisler--perry}, visually indistinguishable conditions, issues one large saccade to the target and stops: the same outcome reached by a different process, and the divergence is in the \emph{order} and \emph{extent} of fixations rather than in their location. Row~6 is a human miss the model does not make. Under gist-$k{=}32$ and \textsc{crop} the model does not search longer but terminates in false absence. Rows are illustrative, selected as trials spanning the range of human search difficulty; quantitative claims on the finding axis rest on all \NImagesTP\ target-present trials.}
  \label{fig:qual}
\end{figure}

Human goal-directed visual search is a \emph{serial} process. The fovea resolves fine detail only within
$\sim$1--2$^{\circ}$ of gaze, so a target detected coarsely in the periphery must be fixated before it can
be confirmed; search unfolds as a sequence of saccades whose targeting, extent and termination are well
characterised, theoretically \cite{heaton2020serial,gupta2022asymmetry}, computationally
\cite{bujia2022bayes,travi2022bench,yang2020irl,yang2024hat}, and empirically on datasets such as
COCO-Search18 for target-present and target-absent search \cite{chen2021cocosearch18,chen2022targetabsent}.
This seriality is imposed by the optics of the eye and is what a foveated agent must pay to search.

This motivates a falsifiable hypothesis: if a model views a scene through the \emph{same} foveated,
acuity-limited aperture as a human (sharp at gaze, degraded in the periphery, displaceable only by
re-fixating) then matched retinal input ought to induce matched search behaviour. The assumption is
load-bearing for two common practices: using multimodal large language models (MLLMs) as stand-ins for
human observers, and reading attention-alignment scores (the overlap between model attention and human
fixations) as evidence that a model ``sees like us''. Both take behavioural or spatial agreement to
license a claim about \emph{process}. We test this by decomposing ``human-like search'' into three axes and
asking on which, if any, an MLLM resembles a human: \emph{decision} (target present or absent?),
\emph{finding} (does gaze reach the target, at what cost?), and \emph{gaze} (are the eye-movement dynamics
human-like?). Three general-purpose MLLMs from three families (Qwen3.5-35B-A3B~\cite{qwen3.5}, GLM-4.6V-Flash~\cite{vteam2025glm45vglm41vthinkingversatilemultimodal},
Gemma-4-E4B~\cite{gemmateam2026gemma4technicalreport}) search COCO-Search18 fixation-by-fixation through a byte-identical foveation harness, against
ten human scanpaths per scene.

\begin{figure}[t]\centering
  \includegraphics[width=\linewidth]{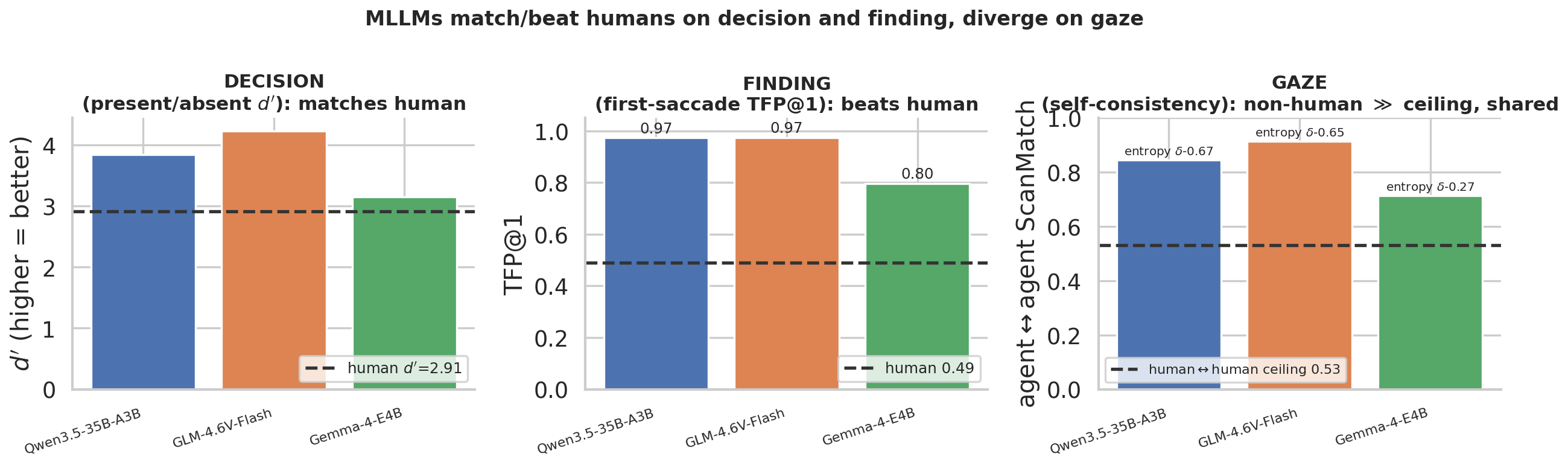}
  \caption{The three-axis dissociation at the human-matched condition. \emph{Decision} ($d'$,
  present/absent): the models match or exceed the human reference. \emph{Finding} (first-saccade TFP@1): the models
  exceed the human rate (\HumanTFPone) by a wide margin. \emph{Gaze} (cross-seed self-consistency): all
  three lie above the human$\leftrightarrow$human ceiling (\HumanCeiling) and apart from the human
  (bar height, cross-seed ScanMatch; annotation, gaze-entropy Cliff's $\delta$), with Gemma-4-E4B nearest. The correct outcome is produced by a shared,
  non-human process.}
  \label{fig:diss}
\end{figure}

The axes \emph{dissociate} (Fig.~\ref{fig:diss}). On decision and finding the models are
human-or-better (near-ceiling present-target detection and first-saccade target fixation
\QwenGPTFPone/\GlmGPTFPone/\GemmaGPTFPone\ versus the human \HumanTFPone, at comparable eventual success and
no more fixations) yet on gaze they are non-human, the three sharing \emph{one} low-entropy,
large-amplitude, highly self-consistent signature that agrees with itself far more than with any human
(cross-seed ScanMatch \QwenGPSelfConsist/\GlmGPSelfConsist/\GemmaGPSelfConsist\ against the inter-observer
ceiling \HumanCeiling). Fig.~\ref{fig:qual} shows what this looks like on individual trials: the model lands on the target while the human is still accumulating fixations, so the two agree on the answer and on roughly where to look while differing in how the looking unfolds. The correct outcome is produced by a shared, non-human process, consistent with a
single-pass, non-serial architecture rather than a limit of acuity, as matched retinal \emph{input}
reproduces \emph{where} humans look but not \emph{how} looking unfolds (\S\ref{sec:gaze}). Our contributions
follow: answer-alignment and saliency metrics are blind to this process axis and cannot certify human-like
vision; zero-shot MLLMs suit outcome and spatial questions but not process and temporal ones; and a
non-serial searcher that matches human outcomes is a null model for the human serial bottleneck. We do not
propose a scanpath predictor or compete on accuracy.

\section{Related work}

\textbf{Human search and its models.} COCO-Search18 provides laboratory human fixations for target-present
and target-absent search \cite{chen2021cocosearch18,chen2022targetabsent} and anchors scanpath
predictors, inverse-reinforcement-learning and transformer
models \cite{mondal2023gazeformer,yang2020irl,yang2024hat},
adversarially and self-supervised trained
variants~\cite{10222686,9857302},
domain-adapted and stochastic generators for out-of-domain
stimuli~\cite{10.1145/3549555.3549597,kerkouri2026spgenstochasticscanpathgeneration}
and ideal-observer/Bayesian searchers benchmarked on
common data \cite{bujia2022bayes,travi2022bench}. Classically, search mixes parallel peripheral evaluation
with serial focal inspection \cite{heaton2020serial,wolfe2021guided}, and search asymmetries emerge from natural-image
statistics rather than task-specific training \cite{gupta2022asymmetry}. These \emph{predict or explain
human} attention; we characterise an MLLM's \emph{process} against the same reference, taking seriality as
the property we test.

\textbf{Foveation as a modelling constraint.} The Geisler--Perry acuity falloff~\cite{geisler1998foveation} supplies our human-matched renderer. Foveated architectures have been studied both as models of human representation,  emergent properties of foveated perceptual  systems~\cite{deza2020emergent}, central--peripheral division in scene recognition~\cite{wang2017central}, and human-like representation under variable resolution~\cite{gizdov2025seeing,gizdov2024variable}, and as search mechanisms, with foveal detectors trained to search directly~\cite{paula2023learning}.
Foveated-observer models capture human performance that non-foveated metrics miss, medical search \cite{lago2021foveated}, foveated transformers \cite{jonna2022foveatedvit}, scene-understanding time \cite{wen2025fsum}, dual ``what''/``where'' saccade selection \cite{dauce2020dual}, and semantically guided foveal models predict human
scanpaths on COCO-Search18~\cite{luzio2024semantic} \cite{luzio2025semba}, as do
architectures for viewing geometries where only part of the scene is
resolvable at once~\cite{Kerkouri2022SalyPath360SA}. Most pointedly, biologically constrained networks viewing scenes foveally produce human-like scanpaths \emph{without being trained to} \cite{neva,nevaclip}. That foveation has repeatedly \emph{sufficed} to induce human-like search motivates our question; for a general-purpose MLLM it does not (Sec.~4.3).

\textbf{MLLM versus human vision.} A growing literature asks whether MLLMs perceive as humans
do, cognitive paradigms \cite{burden2025ispy,lin2024hvsbench,blink,tet}, which find that models ``see but do
not perceive'', and attention comparison via eye-tracking \cite{ghamati2025whichai},\cite{soodMRC},\cite{vqamhug},\cite{10.1145/3797246.3806223} where
attention similarity and task performance dissociate \cite{soodMRC}; critiques of linguistic-prior reliance
separate spatial from semantic guidance \cite{kanade2025doyousee,wu2025indoor}. Closest, MLLM competence
and the human process come apart: a serial deficit inferred from reaction time \cite{budny}, human-trivial
search on which VLMs barely beat chance \cite{berman2025tunnelvision}, correct region with wrong answer
\cite{knowwheretolook}, the mirror of our right-answer/non-human-process finding, but on a static
map, and a capability--strategy gap accuracy benchmarks miss. We differ in measuring the
fixation-by-fixation process, with an explicit stopping decision, on an axis these do not assess.

\textbf{Agentic, search-trained MLLMs.} A parallel line adds active perception to raise accuracy, LLM-guided
search \cite{wu2023vstar}, tree-based zoom \cite{shen2024zoomeye}, reinforcement-learned focusing
\cite{bai2026glanceorgaze,lai2025minio3,li2025insighto3}, and embodied variants \cite{hstar360,mllmsearch};
chain-of-thought can even \emph{degrade} an embodied searcher \cite{vlnmme}.  These optimise \emph{what} is
answered by learning \emph{where} to look. We instead characterise \emph{how} a general-purpose model
explores under a fixed foveation constraint, without search-specific training; trained agents are the most pertinent next comparison (Sec.~6) but out of scope. A related but distinct line allocates compute rather than acuity, reducing or merging visual tokens for efficiency~\cite{bolya2022token,zheng2024enhancing}; these are budget-allocation mechanisms, not models of peripheral vision.

\section{Method}
 \vspace{-3mm}

\begin{figure}[t]
\centering
\begin{tikzpicture}[
  font=\footnotesize, color=loopInk,
  node distance=4.4mm,
  stage/.style={draw, semithick, rounded corners=1.6pt, align=center,
                inner sep=3pt, text width=54mm, minimum height=7mm},
  setup/.style ={stage, draw=loopSlate, fill=loopSlateBg},
  proc/.style  ={stage, draw=loopBlue,  fill=loopBlueBg},
  gen/.style   ={stage, draw=loopAmber, fill=loopAmberBg},
  term/.style  ={draw, semithick, rounded corners=1.6pt, align=center,
                 inner sep=3pt, text width=41mm, minimum height=7mm,
                 draw=loopGreen, fill=loopGreenBg, text=loopInk},
  dir/.style   ={draw, semithick, rounded corners=1.6pt, align=center,
                 inner sep=3pt, text width=45mm, font=\scriptsize,
                 draw=loopRed, fill=loopRedBg, text=loopInk},
  flow/.style={-{Latex[length=1.7mm,width=1.4mm]}, semithick, draw=loopSlate},
  back/.style={-{Latex[length=1.7mm,width=1.4mm]}, semithick, draw=loopRed},
  tag/.style={font=\scriptsize\bfseries, inner sep=0pt},
]
 
\node[setup] (init)
  {scene $1680\times1050$\,px $+$ target cue\\[1pt]
   forced central fixation $g_0=(0.5,0.5)$};
 
\node[proc, below=of init] (render)
  {\textcolor{loopBlue}{\textbf{foveation renderer}} (deterministic)\\[1pt]
   condition $c$ applied at gaze $g_i$;
   downscale to 1024\,px max side};
 
\node[proc, below=of render] (ctx)
  {\textcolor{loopBlue}{\textbf{context}}: prompt $+$ glimpses $0\ldots i$\\[1pt]
   (earlier glimpses retained)};
 
\node[gen, below=of ctx] (llm)
  {\textcolor{loopAmber}{\textbf{MLLM}} --- one free-form generation, $T=0.6$};
 
\node[proc, below=of llm] (parse)
  {\textcolor{loopBlue}{\textbf{readout}}: final line parsed verbatim\\[1pt]
   (preceding reasoning text discarded)};
 
\draw[flow] (init)   -- (render);
\draw[flow] (render) -- (ctx);
\draw[flow] (ctx)    -- (llm);
\draw[flow] (llm)    -- (parse);
 
\coordinate (bus) at ($(parse.south)+(0,-3.6mm)$);
 
\node[term, anchor=north] (stop) at ($(bus)+(-28mm,-4mm)$)
  {\textcolor{loopGreen}{\textsc{found}\,$(x,y)$ \textbf{or} \textsc{absent}}\\[1pt]
   episode terminates};
 
\node[dir, anchor=north] (look) at ($(bus)+(28mm,-4mm)$)
  {\textcolor{loopRed}{\textsc{look}}: \texttt{x=$\langle$0..1$\rangle$, y=$\langle$0..1$\rangle$}\\[2pt]
   $g_{i+1}=(x\!\cdot\!W,\; y\!\cdot\!H)$ px, $W\!=\!1680$, $H\!=\!1050$;\\
   origin top-left, $y$ grows downward};
 
\draw[semithick, draw=loopSlate] (parse.south) -- (bus);
\draw[flow] (bus) -| (stop.north);
\draw[back] (bus) -| (look.north);
 
\draw[back] (look.east) -- ++(4mm,0)
  |- node[pos=0.22, right=0.6mm, font=\scriptsize, text=loopRed, align=left]
     {next gaze\\$g_{i+1}$}
  (render.east);
 
\node[left=3.2mm of llm.west, font=\scriptsize, text=loopSlate,
      align=right, text width=17mm] (cap)
  {50-glimpse cap\\(never binds)};
\draw[dotted, semithick, draw=loopSlate] (cap.east) -- (llm.west);
 
\end{tikzpicture}
\caption{The search loop. Each episode begins at a forced central
fixation; the renderer applies the active foveation condition at the
current gaze point, the resulting glimpse is appended to the context,
and the model returns a single directive line. A \textsc{look} directive
supplies normalised coordinates that are mapped to display pixels and
become the next gaze point, closing the loop (red); \textsc{found} or
\textsc{absent} terminates the episode (green). No search policy is
imposed and the glimpse cap never forces termination: every episode ends
on the model's own decision.}
\label{fig:loop}
\end{figure}
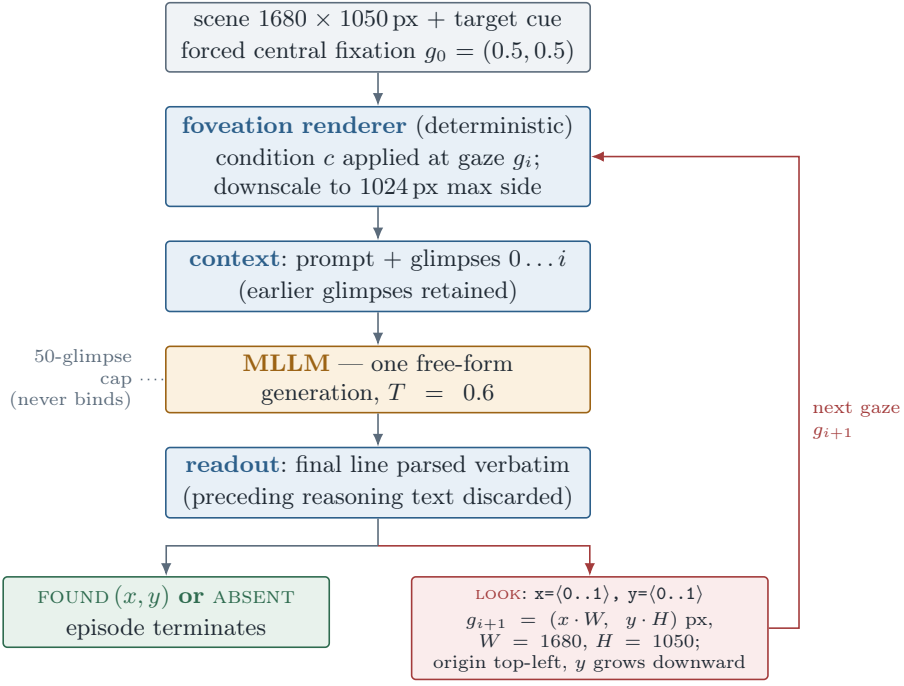
 
\textbf{Data.} The human reference is COCO-Search18~\cite{chen2021cocosearch18}: ten observers per scene
issuing a gamepad present/absent decision on $1680{\times}1050$\,px scenes subtending $\sim$$54°{\times}35°$
($\to\sim$$30$\,px/deg). We use the validation split and a frozen, category-stratified subset of
\NImagesTP\ target-present and \NImagesTA\ target-absent scenes (all ten human scanpaths each); the unit
of analysis is the (scene, target) trial. 

\begin{figure}[!tb]\centering
  \includegraphics[width=0.58\linewidth]{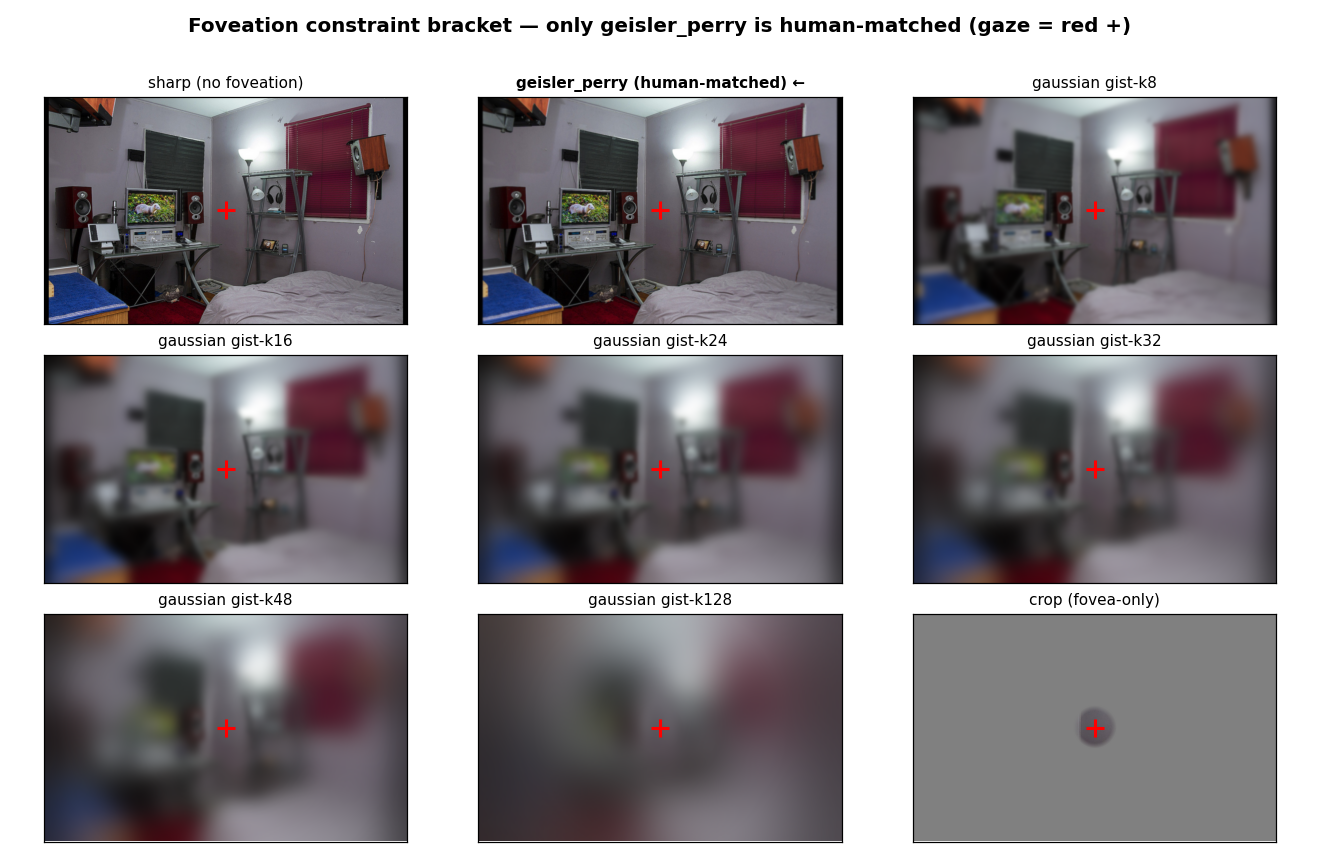}
  \caption{The foveation bracket applied to one scene at a fixed gaze point (red $+$). Only
  \textsc{geisler--perry} is human-matched, and at this geometry it is mild, nearly indistinguishable
  from \textsc{sharp}; the gaussian gist ladder and \textsc{crop} are synthetic anchors.}
  \label{fig:bracket}
\end{figure}

\textbf{Foveation bracket.} At the gaze point a deterministic
renderer applies one of four condition \emph{families} (nine conditions in total; Fig.~\ref{fig:bracket}): \textsc{sharp}; \textsc{geisler--perry}
(GP), the Geisler and Perry~\cite{geisler1998foveation} acuity falloff (the only condition calibrated to human acuity; its mild appearance reflects that peripheral information at this viewing geometry is more legible than intuition suggests, and follows the calibrated falloff of Sec.~S1); \textsc{gaussian} ``gist-$k$''
(``gist'' denotes the coarse peripheral information surviving foveation), the same
falloff with the peripheral cutoff demand scaled by
$k \in \{8, 16, 24, 32, 48, 128\}$, so that GP is the $k{=}1$ member of this
family, run with the eye's measured constants, and the \textsc{gaussian}
conditions are the same equation deliberately detuned (a synthetic degradation not
tuned to human behaviour); and
\textsc{crop}, a fovea-only disc. 

\textbf{Procedure.} As depicted in Fig. \ref{fig:loop}, each episode begins at a forced central fixation; at
every step the model observes the scene rendered at its gaze point (earlier glimpses retained in context)
and returns one directive: \textsc{look}, \textsc{found}, or \textsc{absent}. We use ``scanpath''/``gaze'' for the model's sequence of requested fixation coordinates: an operational analogue of oculomotor scanning, not a claim that the model has eye movements. No search policy is
imposed; the $50$-glimpse cap never forces termination (episodes end on the model's own \textsc{found}/\textsc{absent} decision). Decisions use a single free-form
generation per step at temperature $0.6$ under \NSeeds\ seeds; the harness, prompt (see the supplementary material) and renderer are identical across models. The Geisler--Perry falloff is computed in the native $1680{\times}1050$ frame and each glimpse is then downscaled uniformly to a $1024$-px maximum side, preserving the falloff geometry up to a global scale. 

\textbf{Directive readout.} Each generation ends with a single directive line, parsed verbatim: \texttt{LOOK: x=$\langle$0..1$\rangle$, y=$\langle$0..1$\rangle$}, \texttt{FOUND: x=$\langle$0..1$\rangle$, y=$\langle$0..1$\rangle$}, or \texttt{ABSENT}. Coordinates are normalised to the unit square with the origin at the top-left and $y$ growing downward, a convention fixed in the prompt (Sec.~S12, in supplementary materials); the requested point is mapped to display pixels and becomes the next gaze position. Only the final directive line is consumed, so any preceding reasoning text does not affect the readout. The 50-glimpse cap bounds runaway episodes but never terminates one: every episode ends on the model's own \textsc{found}/\textsc{absent} decision. The first-saccade rate is therefore not an artefact of coordinate parsing: it is corroborated by high eventual success (TFP-end 0.98/0.98/0.93), low median fixation counts (2/2/3), and stability across hit tolerances (max $|\Delta$TFP@1$| \leq 0.095$ over $\pm$0.5/1/1.5$^\circ$, Table~S8).

\textbf{Models.} Qwen3.5-35B-A3B (a 35B mixture-of-experts, 3B active;
reasoning-tuned), GLM-4.6V-Flash ($\approx$$9$B; reasoning-tuned) and Gemma-4-E4B ($\approx$$4$B;
instruction-tuned, run with thinking disabled), which differ in architecture, family and training recipe; as these factors covary,
any cross-model trend is reported descriptively.\footnote{HuggingFace checkpoints \texttt{Qwen/Qwen3.5-35B-A3B}, \texttt{zai-org/GLM-4.6V-Flash}, \texttt{google/gemma-4-E4B-it}. The two reasoning-tuned models run with their default thinking and Gemma-4-E4B with thinking disabled; at each step only the final directive line is consumed.} 

\textbf{Measures.} Search is interpreted only on trials
passing the one-shot existence test, whose detection ceiling is near-identical across models
(Sec.~\ref{sec:dec}). We report, for \emph{decision}, existence accuracy and signal-detection $d'$ (criterion in Sec.~S4); for \emph{finding}, target-fixation probability (TFP) by saccade, with a hit defined as a
fixation within $1°$ of the target box, and fixation count; and for \emph{gaze}, the per-scanpath signature
(saccade amplitude, entropy, refixation, length, center bias, turning angle, coverage) as Cliff's $\delta$
against humans, scanpath
similarity~\cite{cristino2010scanmatch} against the human$\leftrightarrow$human
ceiling, and a cross-model PCA. All measures are defined in Sec.~S3; durations are human-only.
 
  \vspace{-5mm}
\section{Results}
 \vspace{-3mm}
The three models were evaluated on \NImages\ scenes, \NSeeds\ seeds and nine conditions
($3$ models $\times$ \NImages\ scenes $\times$ \NSeeds\ seeds $\times$ nine conditions $=$ \NEpisodes\ episodes), all completed (Sec.~S10). Quantities are reported
at the human-matched GP condition unless a sweep is specified; complete per-condition tables for all three
models appear in the supplement.

  \vspace{-5mm}
\subsection{Decision and finding: models match or exceed humans}
\label{sec:dec}
 \vspace{-3mm}

\begin{table}[!tb]\centering
\caption{Outcome by foveation condition (existence-passed trials); each entry is Qwen/GLM/Gemma (Q/G/Gm) against the human reference (first row; per-model $n$ differs slightly). Finding and outcome measures match or exceed the human values; the \emph{dynamic} gaze divergence is quantified in Table~\ref{tab:gaze}, though target-absent search length (NFix-TA) also differs from humans for Qwen. TFP@$1$/TFP-end: target fixated on the first / by the final saccade (TP); NFix-TP/-TA: median fixations on target-present/absent trials; TA decl.-abs: target-absent declared-absent rate; dens.\ CC: correlation of model and human fixation-density maps.}
\label{tab:outcome}
\setlength{\tabcolsep}{4pt}
\resizebox{\linewidth}{!}{\begin{tabular}{l *{18}{c}}
\toprule
& \multicolumn{3}{c}{TFP@1} & \multicolumn{3}{c}{TFP-end} & \multicolumn{3}{c}{NFix-TP}
& \multicolumn{3}{c}{NFix-TA} & \multicolumn{3}{c}{TA decl.-abs} & \multicolumn{3}{c}{dens.\ CC} \\
\cmidrule(lr){2-4}\cmidrule(lr){5-7}\cmidrule(lr){8-10}\cmidrule(lr){11-13}\cmidrule(lr){14-16}\cmidrule(lr){17-19}
condition & Q & G & Gm & Q & G & Gm & Q & G & Gm & Q & G & Gm & Q & G & Gm & Q & G & Gm \\
\midrule
\textbf{human} & \multicolumn{3}{c}{\textbf{0.49}} & \multicolumn{3}{c}{\textbf{0.93}}
& \multicolumn{3}{c}{\textbf{3}} & \multicolumn{3}{c}{\textbf{5}}
& \multicolumn{3}{c}{---} & \multicolumn{3}{c}{---} \\
\midrule
sharp                  & 0.97 & 0.97 & 0.80 & 0.97 & 0.98 & 0.93 & 2 & 2 & 3 & 1 & 4 & 6 & 0.94 & 0.91 & 0.96 & 0.55 & 0.54 & 0.49 \\
\textbf{GP} & 0.97 & 0.97 & 0.80 & 0.98 & 0.98 & 0.93 & 2 & 2 & 3 & 1 & 4 & 5 & 0.93 & 0.91 & 0.96 & 0.58 & 0.63 & 0.50 \\
k8                     & 0.93 & 0.84 & 0.64 & 0.95 & 0.89 & 0.86 & 3 & 2 & 3 & 2 & 4 & 6 & 0.84 & 0.77 & 0.81 & 0.65 & 0.45 & 0.26 \\
k16                    & 0.78 & 0.71 & 0.42 & 0.91 & 0.80 & 0.75 & 3 & 3 & 4 & 4 & 4 & 6 & 0.78 & 0.77 & 0.68 & 0.70 & 0.38 & 0.45 \\
k24                    & 0.56 & 0.49 & 0.28 & 0.81 & 0.65 & 0.65 & 3 & 3 & 6 & 6 & 4 & 6 & 0.72 & 0.68 & 0.58 & 0.74 & 0.46 & 0.29 \\
k32                    & 0.42 & 0.33 & 0.19 & 0.71 & 0.53 & 0.59 & 3 & 3 & 6 & 7 & 4.5 & 7 & 0.62 & 0.65 & 0.52 & 0.70 & 0.43 & 0.26 \\
k48                    & 0.28 & 0.18 & 0.10 & 0.56 & 0.36 & 0.47 & 5 & 4 & 6 & 7 & 4 & 7 & 0.46 & 0.65 & 0.54 & 0.49 & 0.33 & 0.25 \\
k128                   & 0.15 & 0.09 & 0.01 & 0.31 & 0.22 & 0.39 & 6 & 4 & 6 & 6 & 4 & 6 & 0.85 & 0.86 & 0.78 & 0.18 & 0.14 & 0.19 \\
crop                   & 0.04 & 0.04 & 0.00 & 0.20 & 0.11 & 0.16 & 3 & 3 & 5 & 2 & 3 & 5 & 0.68 & 0.50 & 0.67 & 0.25 & 0.11 & 0.15 \\
\bottomrule
\end{tabular}}
 \vspace{-2mm}
\end{table}
\textbf{Decision.} One-shot present-target detection is near-ceiling and comparable across models (target-present /
target-absent existence accuracy \QwenTPexist/\QwenTAexist, \GlmTPexist/\GlmTAexist,
\GemmaTPexist/\GemmaTAexist; false-positive bias on absent scenes
\QwenYesBias/\GlmYesBias/\GemmaYesBias), so subsequent search divergences cannot be attributed to
detection failure and the existence-passed sets are comparable. In the agentic task the present/absent
decision is highly sensitive for every model ($d'$ \QwenGPDprime/\GlmGPDprime/\GemmaGPDprime, against the
human \HumanDprime; Sec.~S4), and absent scenes are declared absent at \QwenGPDeclAbs/\GlmGPDeclAbs/\GemmaGPDeclAbs\ --- distinct from the one-shot existence accuracy ($\sim$$0.80$) reported above. \textbf{Finding.} Every model fixates the target on the
first saccade far more often than humans (TFP@1 \QwenGPTFPone/\GlmGPTFPone/\GemmaGPTFPone\ against
\HumanTFPone; Table~\ref{tab:outcome}) and attains comparable eventual success (TFP-end
\QwenGPTFPend/\GlmGPTFPend/\GemmaGPTFPend\ against \HumanTFPend), while issuing no more fixations than
humans (median \QwenGPNumFixTP/\GlmGPNumFixTP/\GemmaGPNumFixTP\ against \HumanNumFixTP). On the two axes
of principal practical interest, whether the decision is correct and whether the target is found, the
models are therefore human-or-better; assessed at the level of outcomes alone, they would be judged
human-like. One target-absent behaviour is, however, already non-human on the search axis: Qwen3.5-35B-A3B declares absence after a single fixation (NFix-TA \QwenGPNumFixTA) versus the human median \HumanNumFixTA, an efficiency win that is itself not human-like.

\vspace{-5mm}
\subsection{Gaze dynamics: a shared, non-human signature}
\label{sec:gaze}
 \vspace{-1mm}
Under the conditions in which outcomes coincide with humans, the eye-movement \emph{process} does not, and
the deviation has the same direction for all three models (Table~\ref{tab:gaze}, Fig.~\ref{fig:gistpca}b; the pattern is visible on single trials in Fig.~\ref{fig:qual}).
Gaze entropy lies below the human value (Cliff's $\delta$ \QwenGPEntropyD/\GlmGPEntropyD/\GemmaGPEntropyD),
indicating spatially concentrated sampling, and saccade amplitudes exceed it
(\QwenGPSaccD/\GlmGPSaccD/\GemmaGPSaccD), indicating direct movements to the target. The scanpaths are
moreover highly self-consistent: cross-seed (agent$\leftrightarrow$agent) ScanMatch is
\QwenGPSelfConsist/\GlmGPSelfConsist/\GemmaGPSelfConsist, far above the human$\leftrightarrow$human
agreement ceiling (\HumanCeiling), so each model reproduces its own scanpath more closely than two humans
agree. A principal-component analysis of the five-statistic signature places all three models, under the
legible conditions, in a region disjoint from the human reference (Fig.~\ref{fig:gistpca}b). That three
unrelated models occupy the \emph{same} region is the multivariate expression of the gaze divergence. These effects survive per-metric mixed-effects models with crossed scene and rater random intercepts (gaze entropy and saccade amplitude have $95\%$ intervals excluding zero for all three models; Sec.~S9).
 

\begin{table}[t]
\caption{Gaze axis at the human-matched condition and at two intermediate
degradations, re-analysed from the per-condition tables of the submission
(Tables~S3--S4). Cliff's $\delta$ against the human distribution ($|\delta|>0.33$
in bold; sign is agent $-$ human) and cross-seed self-consistency. At GP the
deviation is shared: low entropy, large amplitudes, near-human center bias. At
k16--k24 the amplitude effect vanishes, refixation rises, and Gemma-4-E4B's
entropy reverses sign, so the GP signature is a property of the legible regime
rather than a fixed offset; self-consistency above the ceiling is what persists
for the two reasoning-tuned models.}
\label{tab:gaze}
\centering\small
\begin{tabular}{llccccc}
\toprule
cond. & model & entropy $\delta$ & sacc.\ amp $\delta$ & refix.\ $\delta$ & center bias $\delta$ & agent$\leftrightarrow$agent \\
\midrule
\multicolumn{2}{l}{human$\leftrightarrow$human ceiling} & --- & --- & --- & --- & 0.53 \\
\midrule
GP  & Qwen3.5-35B-A3B & \textbf{$-$.67} & \textbf{$+$.50} & $+$.04 & $-$.18 & 0.84 \\
    & GLM-4.6V-Flash  & \textbf{$-$.65} & \textbf{$+$.61} & $-$.13 & $-$.31 & 0.91 \\
    & Gemma-4-E4B     & $-$.27 & $+$.23 & $+$.13 & $+$.06 & 0.71 \\
\midrule
k16 & Qwen3.5-35B-A3B & \textbf{$-$.56} & $-$.03 & \textbf{$+$.49} & $+$.01 & 0.78 \\
    & GLM-4.6V-Flash  & \textbf{$-$.53} & $+$.20 & $+$.30 & $-$.05 & 0.80 \\
    & Gemma-4-E4B     & $+$.19 & $+$.07 & $+$.27 & $+$.24 & 0.59 \\
\midrule
k24 & Qwen3.5-35B-A3B & \textbf{$-$.33} & $-$.08 & \textbf{$+$.52} & $+$.02 & 0.70 \\
    & GLM-4.6V-Flash  & \textbf{$-$.41} & $+$.02 & \textbf{$+$.45} & $-$.02 & 0.71 \\
    & Gemma-4-E4B     & \textbf{$+$.36} & $+$.16 & \textbf{$+$.38} & $+$.19 & 0.48 \\
\bottomrule
\end{tabular}
\end{table}

\subsection{No regime recovers human-like search; a single-pass account}

\textbf{Where the gaze gap is narrowest, the models are failing rather than searching.} At GP the models operate far above the human first-saccade rate, so the gaze comparison is made at unequal task difficulty; the intermediate conditions k16 and k24 bring TFP@1 to 0.78/0.71/0.42 and 0.56/0.49/0.28 against the human 0.49, and there the gaze signature partly converges (Table~\ref{tab:gaze}). The convergence is not evidence of human-like search. At k24 eventual success has already fallen to 0.81/0.65/0.65 against the human 0.93, so the models are not matching the human process at matched difficulty but failing to resolve the scene; what rises with the convergence is refixation ($\delta$ $+$.52/$+$.45/$+$.38, against $+$.04/$-$.13/$+$.13 at GP), the revisiting of cells the model cannot resolve, which is the failure signature of Sec.~S8 rather than the inspection-driven refixation of a serial searcher. The argument of Sec.~4.1 therefore extends from outcomes to gaze: there is no operating point at which a
model is human-like on both axes at once. Across the legible range what persists
is determinism (cross-seed ScanMatch 0.78/0.80 at k16 and 0.70/0.71 at k24 for the two reasoning-tuned models, against the 0.53 ceiling) while Gemma-4-E4B falls to 0.48 at k24, below the ceiling, and is the exception.

\begin{figure}[!tb]
  \begin{subfigure}{0.5\linewidth}\centering
    \includegraphics[width=0.80\linewidth]{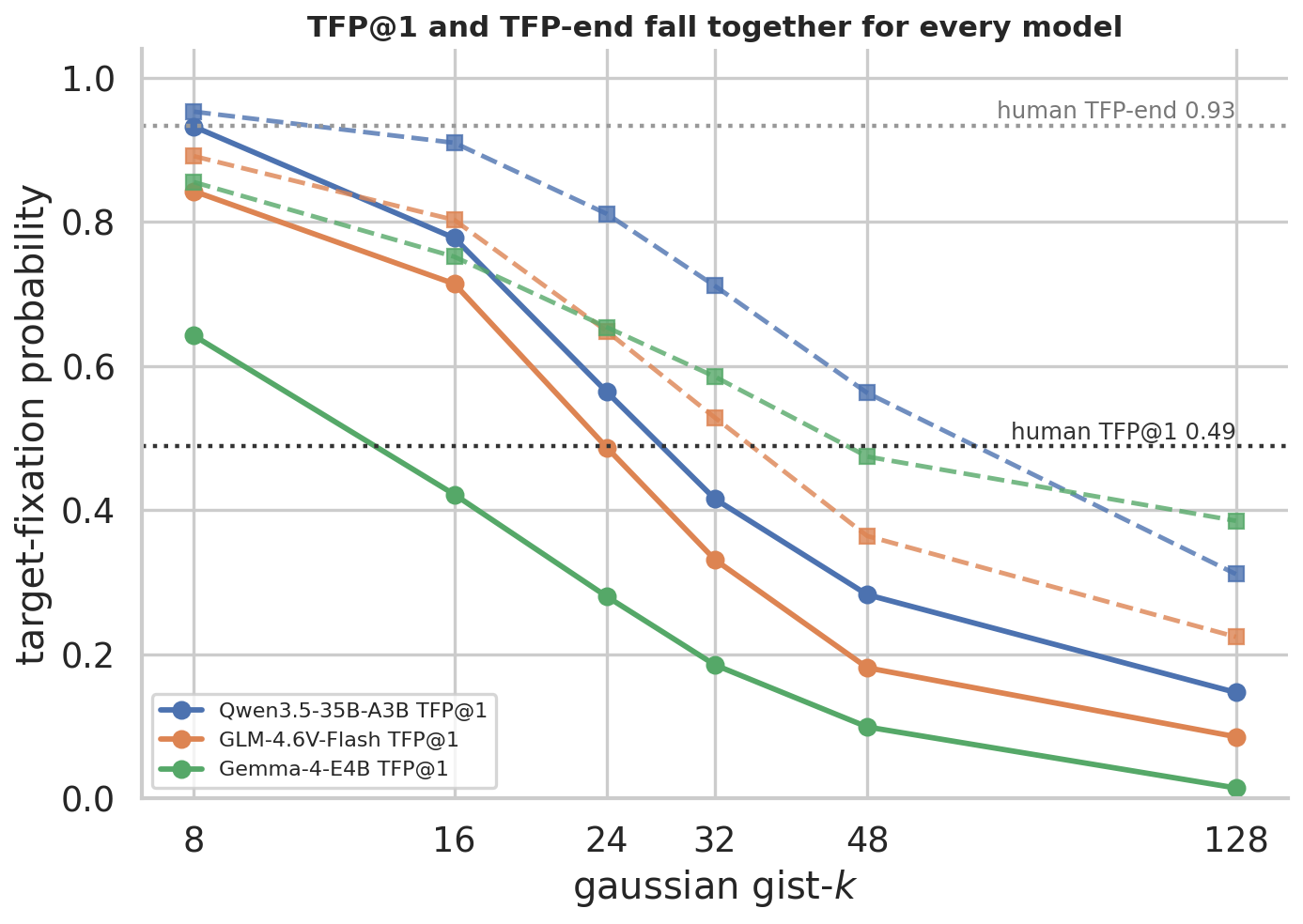}\caption{No human-like regime.}\label{fig:gist}
  \end{subfigure}\hfill
  \begin{subfigure}{0.49\linewidth}\centering
    \includegraphics[width=0.80\linewidth]{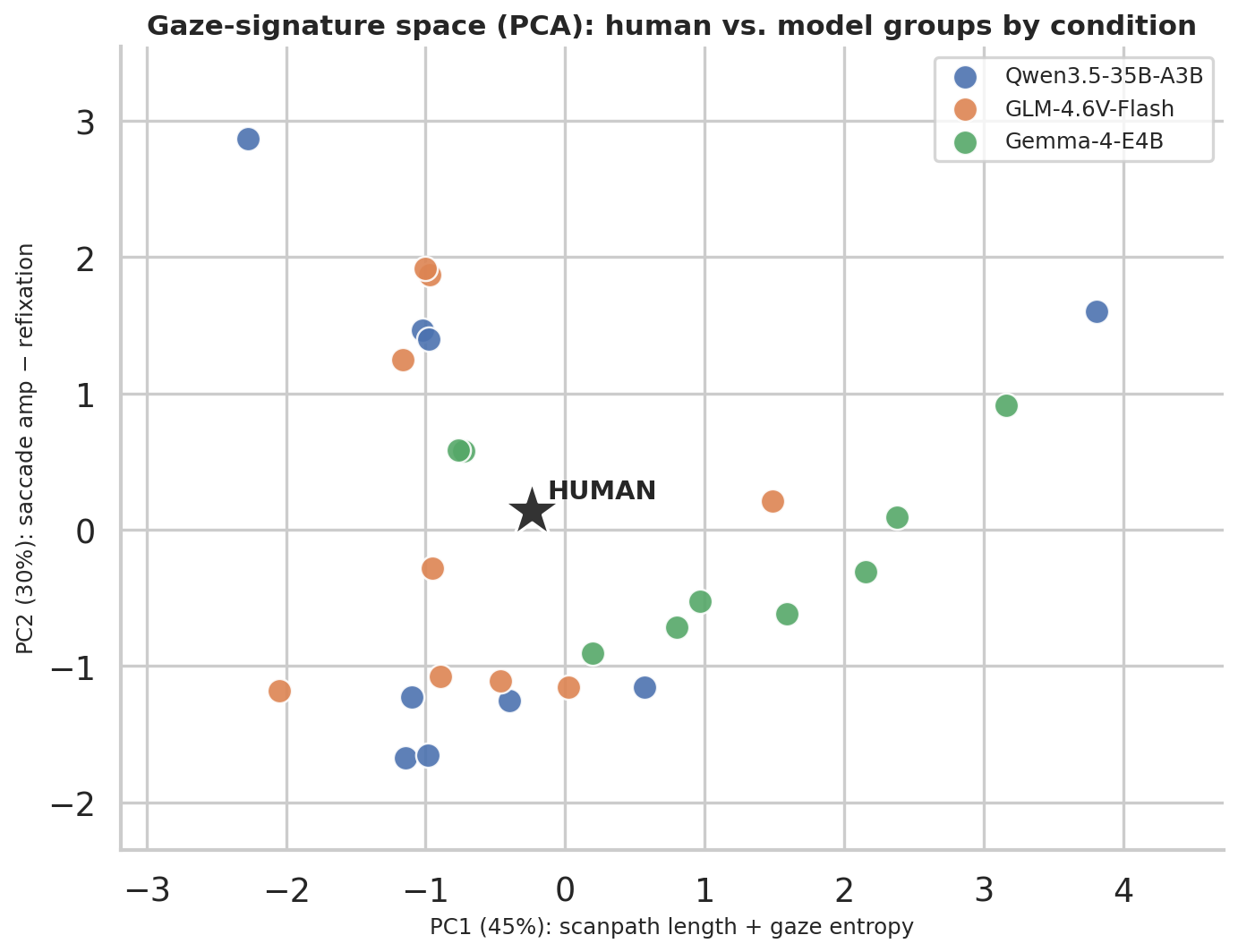}\caption{Gaze-signature PCA (per group).}\label{fig:pca}
  \end{subfigure}
  \caption{(\subref{fig:gist}) First-saccade targeting (solid) and eventual success (dashed) as functions
  of the degradation factor $k$ for the three models, with human reference levels; the two quantities fall
  together, so no $k$ yields human-like search at human-like success. (\subref{fig:pca}) Principal-component
  analysis of the per-(model,\,condition) five-statistic gaze signature (PC1, \PCApcone\%: scanpath length
  and gaze entropy; PC2, \PCApctwo\%: saccade amplitude versus refixation). Under the legible conditions each
  model sits in a region offset from the human reference ($\star$); points approach the human only under
  severe degradation, with Gemma-4-E4B nearest.}
  \label{fig:gistpca}
  \vspace{-7mm}
\end{figure}

Sweeping the synthetic degradation does not yield a level at which a model searches like a human (TFP@1
$\approx$ \HumanTFPone) while still locating the target (TFP-end $\approx$ \HumanTFPend). For none of the
three models (Fig.~\ref{fig:gistpca}a) do these quantities separate: they decline together, so by the
degradation that lowers first-saccade targeting to the human rate ($k{=}32$ for Qwen3.5-35B-A3B,
TFP@1/TFP-end \QwenKthirtytwoTFPone/\QwenKthirtytwoTFPend; $k{=}16$ for Gemma-4-E4B,
\GemmaKsixteenTFPone/\GemmaKsixteenTFPend), eventual success has already collapsed. The divergence is
not an effect of acuity: the human-matched condition is behaviourally indistinguishable from \textsc{sharp}
for every model (e.g.\ TFP@1 \QwenGPTFPone\ against \QwenSharpTFPone; Tables~\ref{tab:outcome}--\ref{tab:gaze}),
even though it measurably degrades the periphery. Nor is it a difference of spatial prior: center bias is human-like for all three (sub-threshold $\delta$, Table~\ref{tab:gaze}), so the models look in broadly human-relevant places, whereas the order and dynamics of their fixations (low entropy, large saccades, high determinism) do not. The missing component is serial sampling, not the spatial prior; we do not manipulate architecture directly, so this attribution is by elimination (acuity and spatial prior ruled out) and a causal test imposing serial sampling is left to future work. As evidence is removed the models do not lengthen search gracefully: refixation rises, then at the most severe degradation the declared-absent rate rebounds as they default to absent, a model-internal failure signature needing no human reference (Fig.~\ref{fig:qual}, gist-$k{=}32$ and \textsc{crop} columns; Sec.~S8).
 
 \vspace{-5mm}
\subsection{Generality and robustness}
The dissociation and the shared gaze signature hold across three distinct model families, suggesting a
property of the foveated-MLLM paradigm rather than of a single instance. The
first-saccade advantage holds within every eccentricity-by-size difficulty stratum (Sec.~S9) and is
insensitive to the target-box tolerance: across tolerances of $0.5$, $1$ and $1.5°$, TFP@1 shifts by at
most $\sim$$0.1$. The high cross-seed determinism is not an artifact of low-temperature decoding; in a
temperature sweep on our anchor model (Qwen3.5-35B-A3B) it remains well above the inter-observer ceiling even at
temperature $1.0$ on the legible conditions (Sec.~S9). Cross-seed and human inter-observer agreement are not identical constructs (no matched human intra-observer baseline exists in COCO-Search18), so we treat the determinism gap as suggestive, resting it on this temperature-$1.0$ persistence. The \emph{magnitude} of the gaze deviation follows a consistent ordering across models: Gemma-4-E4B is the most human-like model, with the smallest effect on five of the seven statistics (entropy, saccade amplitude, scanpath length, center bias, coverage; e.g.\ gaze-entropy $\delta$ \GemmaGPEntropyD\ vs.\ Qwen \QwenGPEntropyD; cross-seed ScanMatch \GemmaGPSelfConsist\ vs.\ \QwenGPSelfConsist), and its number of fixations (search extent) is the only fixed effect whose mixed-effects $95\%$ interval contains zero (Sec.~S9). Because architecture, recipe and sparsity covary across three models, we read this as weaker single-pass targeting, not a more human-like strategy, and report it descriptively.
 
 \vspace{-5mm}
\section{Discussion}
\paragraph{Metrics and surrogacy.} The models match the present/absent answer and partially match the
human saliency map (density correlation \QwenGPDensityCC/\GlmGPDensityCC/\GemmaGPDensityCC) yet diverge on
every temporal measure of gaze. Because answer-alignment and saliency scores are computed on outcomes or a
time-collapsed map, a high value is necessary but not sufficient: zero-shot MLLMs are adequate surrogates
for \emph{outcome and spatial} questions but not \emph{process and temporal} ones, a property of the class
that model selection does not remove.
 
\paragraph{A non-human searcher as a null model.} Conversely, a system attaining human-or-better outcomes
\emph{without} a serial bottleneck is a useful null: it isolates the behaviours due to seriality from the detection and spatial priors it
already reproduces.

\vspace{-5mm}
\section{Limitations and scope}
\label{sec:lim}
Our scope is general-purpose models under a fixed foveation constraint: search-trained agentic,
pointing-native and frontier closed-source models---the most pertinent next comparison---are not evaluated.
The cross-model trend is confounded (architecture, recipe, sparsity) and reported descriptively; the gaze
signature is measured under a single prompt whose memory clause may partly shape refixation; and fixation
durations are human-only.
 
\section{Conclusion}
Three MLLMs, driven fixation by fixation through a foveation calibrated
to human acuity, match or exceed humans on the decision and on target
acquisition. Their gaze does not follow: at the legible conditions all
three share one low-entropy, large-amplitude, highly self-consistent
signature, and no degradation regime recovers human-like search at
human-like success, where the signature converges the models are
failing to resolve the scene rather than searching. Matched retinal
input therefore reproduces \emph{where} humans look but not \emph{how}
the looking unfolds, a dissociation consistent with a single-pass reader
carrying a human-like spatial prior rather than with a limit of acuity.
Because answer-alignment and saliency scores are computed on outcomes or
on a time-collapsed map, they cannot certify correspondence on this
\emph{process} axis; conversely, a searcher attaining human-or-better
outcomes without a serial bottleneck is a null model against which the
behavioural cost of seriality can be isolated.

\bibliographystyle{splncs04}
\bibliography{main}

\clearpage
\setcounter{section}{0}\setcounter{table}{0}\setcounter{figure}{0}\setcounter{equation}{0}
\renewcommand{\thesection}{S\arabic{section}}
\renewcommand{\thetable}{S\arabic{table}}
\renewcommand{\thefigure}{S\arabic{figure}}
\renewcommand{\theequation}{S\arabic{equation}}
\begin{center}
  {\LARGE\bfseries Supplementary Material\par}
  \vspace{0.6em}
  {\large Matched Outcomes, Divergent Gaze:\ How Foveated MLLMs Search Compared to Humans\par}
\end{center}
\vspace{1.2em}

This document specifies the experimental methods in full, gives formal definitions of every behavioural
measure and the statistical procedures used to analyse them, and reports the complete per-model results
summarised in the main paper. We first describe the stimuli and the foveation model
(Sec.~\ref{sec:fov}) and the search procedure (Sec.~\ref{sec:proc}); we then define all metrics and the
inferential methodology (Sec.~\ref{sec:metrics}). The three behavioural axes (decision, finding, and
gaze) are reported in Secs.~\ref{sec:decision}--\ref{sec:gaze-supp}, followed by the multivariate analysis
(Sec.~\ref{sec:pca}), the mechanistic analyses (Sec.~\ref{sec:mech}), robustness and cross-model
variation (Sec.~\ref{sec:robust}), data completeness (Sec.~\ref{sec:integrity}), the implications
and scope of the study (Sec.~\ref{sec:scope}), and the prompt (Sec.~\ref{sec:prompt}). All analyses use three vision-language models,
Qwen3.5-35B-A3B (a 35B-parameter mixture-of-experts with 3B active parameters; reasoning-tuned),
GLM-4.6V-Flash ($\approx$9B; reasoning-tuned) and Gemma-4-E4B ($\approx$4B; instruction-tuned), together with the
human reference, on a frozen, category-stratified COCO-Search18 subset of \NImagesTP\ target-present and
\NImagesTA\ target-absent scenes, with ten human scanpaths per scene and \NSeeds\ model seeds per scene
and condition.

\section{Stimuli and the foveation model}\label{sec:fov}
Stimuli, target categories and human scanpaths are drawn from COCO-Search18~\cite{chen2021cocosearch18},
in which observers searched $1680{\times}1050$\,px scenes subtending $\sim$$54°{\times}35°$ of visual
angle for a cued object and reported its presence or absence; the corresponding angular resolution is
$\rho \approx 30$\,px/deg. Visual angle is obtained throughout by $\theta = d/\rho$ for a pixel distance
$d$. The unit of analysis is the \emph{trial} $t=(\text{scene},\text{target})$.
 
Foveation is imposed by a deterministic renderer applied at the current gaze point $g$. Following Geisler
and Perry~\cite{geisler1998foveation}, the highest spatial frequency resolvable by the eye at retinal
eccentricity $e$ (in degrees) is
\begin{equation}
f_c(e) \;=\; \frac{e_2\,\ln(1/CT_0)}{\alpha\,(e+e_2)}\quad\text{cyc/deg},\qquad
e_2{=}2.3°,\ \alpha{=}0.106,\ CT_0{=}1/64 .
\label{eq:gp}
\end{equation}
A Gaussian image pyramid is constructed and, at every pixel of eccentricity $e$, the canonical level
$L=\log_2\!\big(\text{local Nyquist}/f_c(e)\big)$ is selected so that the local image cutoff equals the
eye's, yielding a sharp centre and a smooth peripheral falloff. The four bracket conditions are
(i)~\textsc{sharp}, no foveation; (ii)~\textsc{geisler--perry} (GP), Eq.~\eqref{eq:gp} at $\rho{=}30$,
viewing distance $0.6$\,m, the \emph{only} human-matched condition, and mild at this geometry;
(iii)~\textsc{gaussian} ``gist-$k$'' (\emph{gist}: the coarse, low-resolution peripheral information that survives foveation), identical in form but with the peripheral cutoff demand scaled by a
factor $k\in\{8,16,24,32,48,128\}$, a synthetic degradation \emph{not} tuned to human behaviour; and
(iv)~\textsc{crop}, a fovea-only disc of radius $\sim$$2.5°$ with the periphery removed. The renderer is
identical across models (main Fig.~4 shows the bracket on one scene).

\section{Search procedure}\label{sec:proc}
Each episode begins at a forced central fixation. At step $i$ the model receives the scene rendered under
the active foveation condition at its current gaze point (earlier glimpses retained in context) and must
return exactly one directive: $\textsc{look}(x,y)$, to move the gaze to a new point and continue;
$\textsc{found}(x,y)$, to terminate with a present decision at $(x,y)$; or $\textsc{absent}$, to terminate
with an absent decision. The procedure imposes no search policy: the choice of where to look and when to
stop is the model's alone, and the episode is never terminated on a target hit. A uniform cap of $50$
glimpses bounds runaway episodes. Each glimpse is rendered at the human display resolution and downscaled
to a $1024$-px maximum side before presentation, identically across conditions and models. Decisions are
produced by a single free-form generation per step at sampling temperature $0.6$; each (scene,
condition) is searched under \NSeeds\ independent seeds. The harness, prompt and renderer are byte-for-byte
identical across the three models, so that any behavioural difference is attributable to the model rather
than to the protocol.

\section{Metrics and statistical methodology}\label{sec:metrics}
 
\paragraph{Notation.} A \emph{scanpath} is an ordered sequence of fixations
$s=(f_0,f_1,\dots,f_{L})$ with $f_i=(x_i,y_i)$ in display pixels and $f_0$ the central start; its length
in fixations is $|s|=L{+}1$. For a target-present trial, $B$ denotes the target bounding box and
$B^{\oplus\tau}$ its dilation by a tolerance $\tau$ (default $\tau{=}1°{=}\rho$\,px). A fixation
\emph{hits} the target if $f_i\in B^{\oplus\tau}$, and the first-hit index is
$h(s)=\min\{i: f_i\in B^{\oplus\tau}\}$ (with $h(s)=\infty$ if the target is never fixated). For each
trial we have up to \NSeeds\ model scanpaths (one per seed) and ten human scanpaths; the
\emph{existence-passed} set restricts analysis to trials a model answered correctly on the one-shot
detection test (defined below), applied identically to every model.
 
\subsection{Detection and decision}\label{sec:m-dec}
The one-shot \emph{existence} test presents the full sharp scene with the question ``is there a
\{target\}?''. Existence accuracy is the proportion of correct yes/no answers (separately for
target-present and target-absent scenes), and the \emph{yes-bias} is the false-positive rate on
target-absent scenes, $\Pr(\text{answer}{=}\text{yes}\mid\text{absent})$. The agentic present/absent
decision is summarised by signal-detection theory. With hit rate $H=\Pr(\textsc{found}\mid\text{TP})$ and
false-alarm rate $F=\Pr(\textsc{found}\mid\text{TA})$, and the log-linear correction
$H'=(n_H{+}0.5)/(n_{\text{TP}}{+}1)$, $F'=(n_F{+}0.5)/(n_{\text{TA}}{+}1)$, sensitivity and bias are
\begin{equation}
d' = \Phi^{-1}(H') - \Phi^{-1}(F'), \qquad
c  = -\tfrac{1}{2}\big[\Phi^{-1}(H') + \Phi^{-1}(F')\big],
\label{eq:sdt}
\end{equation}
where $\Phi^{-1}$ is the inverse standard-normal CDF. The human reference uses the recorded gamepad
present/absent responses.
 
\subsection{Finding (efficiency of target acquisition)}\label{sec:m-find}
The \emph{target-fixation probability} (TFP) curve gives, by saccade $n$, the probability that gaze has
reached the target. Over the existence-passed trial set $\mathcal T$,
\begin{equation}
\mathrm{TFP}_n \;=\; \frac{1}{|\mathcal T|}\sum_{t\in\mathcal T}
   \frac{1}{|\mathcal S_t|}\sum_{s\in\mathcal S_t}\mathbf 1\!\big[h(s)\le n\big],
\label{eq:tfp}
\end{equation}
with $\mathcal S_t$ the scanpaths of trial $t$. $\mathrm{TFP@1}$ counts the first saccade after the forced central fixation $f_0$, applied identically to models and humans. We report first-saccade targeting $\mathrm{TFP@1}$ and
eventual success $\mathrm{TFP\text{-}end}=\mathrm{TFP}_{15}$. The same definition is used for the
strata and hit-tolerance analyses, so that a single estimator underlies every TFP value in the paper.
Search extent is the number of fixations per episode, $\mathrm{NumFix}=|s|$, reported as the per-group
median.
 
\subsection{Stopping (target-absent termination)}\label{sec:m-stop}
On target-absent scenes we report the \emph{declared-absent} rate --- the fraction of episodes terminated
with $\textsc{absent}$ --- and the median number of fixations preceding the decision.
 
\subsection{Gaze dynamics (intrinsic scanpath signature)}\label{sec:m-gaze}
The following per-scanpath statistics characterise the eye-movement process independently of task
outcome; all lengths are in degrees of visual angle. The $i$-th saccade has amplitude
$a_i=\lVert f_i-f_{i-1}\rVert/\rho$ and direction $\phi_i=\operatorname{atan2}(y_i-y_{i-1},\,x_i-x_{i-1})$;
the turning angle between successive saccades is $\Delta\phi_i=\operatorname{wrap}(\phi_{i+1}-\phi_i)\in(-180°,180°]$.
Scanpath length is $\ell=\sum_i a_i$; center bias is the mean fixation eccentricity
$\tfrac{1}{|s|}\sum_i\lVert f_i-c\rVert/\rho$ about the scene centre $c$; convex-hull coverage is the area
(deg$^2$) of the convex hull of $\{f_i\}$. Spatial dispersion is quantified by gaze entropy: the display
is partitioned into a $14{\times}9$ grid, and with $p_g$ the fraction of fixations in cell $g$,
\begin{equation}
\mathrm{GazeEntropy} \;=\; -\sum_{g} p_g\,\log_2 p_g \quad\text{(bits)}.
\label{eq:entropy}
\end{equation}
The refixation rate is the fraction of fixations that land in an already-visited grid cell, an
inhibition-of-return signature. Each statistic is summarised by its per-group median and by the effect
size against the human distribution (Sec.~\ref{sec:m-stat}).
 
\subsection{Scanpath similarity}\label{sec:m-sim}
Pairwise scanpath similarity uses ScanMatch~\cite{cristino2010scanmatch}; we retain
it for comparability with the published human$\leftrightarrow$human ceiling,
noting that grid quantisation discards foveal
scale~\cite{10.1145/3797246.3805860} and that similarity can also be scored
semantically rather than
geometrically~\cite{10.1145/3797246.3806223}. Fixations are quantised to the
$14{\times}9$ grid and aligned by Needleman--Wunsch global alignment with a substitution score that
decreases linearly with inter-cell Euclidean distance (threshold $3.5$, gap penalty $0$), normalised by
the maximal self-alignment score to the unit interval (higher is more similar). ScanMatch is computed
in three modes: \emph{agent}$\leftrightarrow$\emph{human} (each model scanpath against each human scanpath
of the trial), \emph{agent}$\leftrightarrow$\emph{agent} (cross-seed model pairs, a measure of
determinism), and \emph{human}$\leftrightarrow$\emph{human} (all $\binom{10}{2}$ human pairs per scene),
the last of which is the inter-observer agreement \emph{ceiling}, \HumanCeiling\ for ScanMatch.
 
\subsection{Fixation-density agreement}\label{sec:m-dens}
A continuous fixation-density map is formed by kernel density estimation,
$\hat D(\mathbf u)\propto\sum_i \mathcal N(\mathbf u;f_i,\sigma^2 I)$ with $\sigma=\rho$ (the central
fixation is excluded). Agreement with the human map is reported as the linear correlation coefficient CC
(Pearson correlation of the two maps), the normalised scanpath saliency NSS (the mean of the $z$-scored
model map sampled at human fixation locations), and the Kullback--Leibler divergence
$\mathrm{KL}=\sum_{\mathbf u} D_h\log(D_h/D_m)$ of the human map from the model map.
 
\subsection{Statistical methodology}\label{sec:m-stat}
Effect sizes against the human distribution use Cliff's $\delta$, the nonparametric dominance statistic
\begin{equation}
\delta(A,B)=\frac{\#\{a>b\}-\#\{a<b\}}{|A|\,|B|}\in[-1,1],
\label{eq:cliffs}
\end{equation}
with sign convention agent $-$ human; $|\delta|>0.33$ is treated as non-trivial. Distributional effects are
confirmed by linear mixed-effects models with crossed random intercepts for scene and rater (human
subject or model seed),
\begin{equation}
y_{ijk}=\beta_0+\textstyle\sum_{c}\beta_c\,\mathbf 1[\text{cond}_{j}{=}c]+u_i+v_k+\varepsilon_{ijk},
\qquad u_i,v_k,\varepsilon \stackrel{}{\sim}\text{indep.\ Gaussian},
\label{eq:lmm}
\end{equation}
where $i$ indexes scene, $k$ rater, $j$ scanpath, and the human condition is the reference level; we
report the fixed effect $\hat\beta_c$ (agent$-$human) and its 95\% confidence interval, treating a CI
that excludes zero as significant. Finally, the joint structure of five of the seven gaze statistics --- gaze entropy, saccade amplitude, refixation, scanpath length and center bias (coverage is dropped from the PCA as collinear with scanpath length and entropy, and turning-angle effects are inconsistent in sign across conditions, so both enter Table~\ref{tab:signature} only) --- is
summarised by a principal-component analysis of the standardised per-group median vectors, giving two
interpretable axes (loadings in Table~\ref{tab:pca}).

\section{Decision axis}\label{sec:decision}
\begin{table}[ht]\centering
\caption{Detection and decision measures. One-shot existence accuracy and yes-bias (full sharp scene),
the in-harness present/absent sensitivity $d'$ and criterion $c$ at the human-matched condition
(Eq.~\eqref{eq:sdt}).}
\label{tab:decision}
\begin{tabular}{lccccc}
\toprule
model & TP exist. & TA exist. & yes-bias & $d'$ (GP) & criterion \\
\midrule
human & --- & --- & --- & 2.91 & 0.03 \\
Qwen3.5-35B-A3B & 0.99 & 0.81 & 0.19 & 3.84 & -0.41 \\
GLM-4.6V-Flash & 0.99 & 0.81 & 0.19 & 4.23 & -0.24 \\
Gemma-4-E4B & 0.97 & 0.80 & 0.20 & 3.14 & 0.38 \\
\bottomrule
\end{tabular}

\end{table}
\begin{figure}[ht]\centering\includegraphics[width=0.84\linewidth]{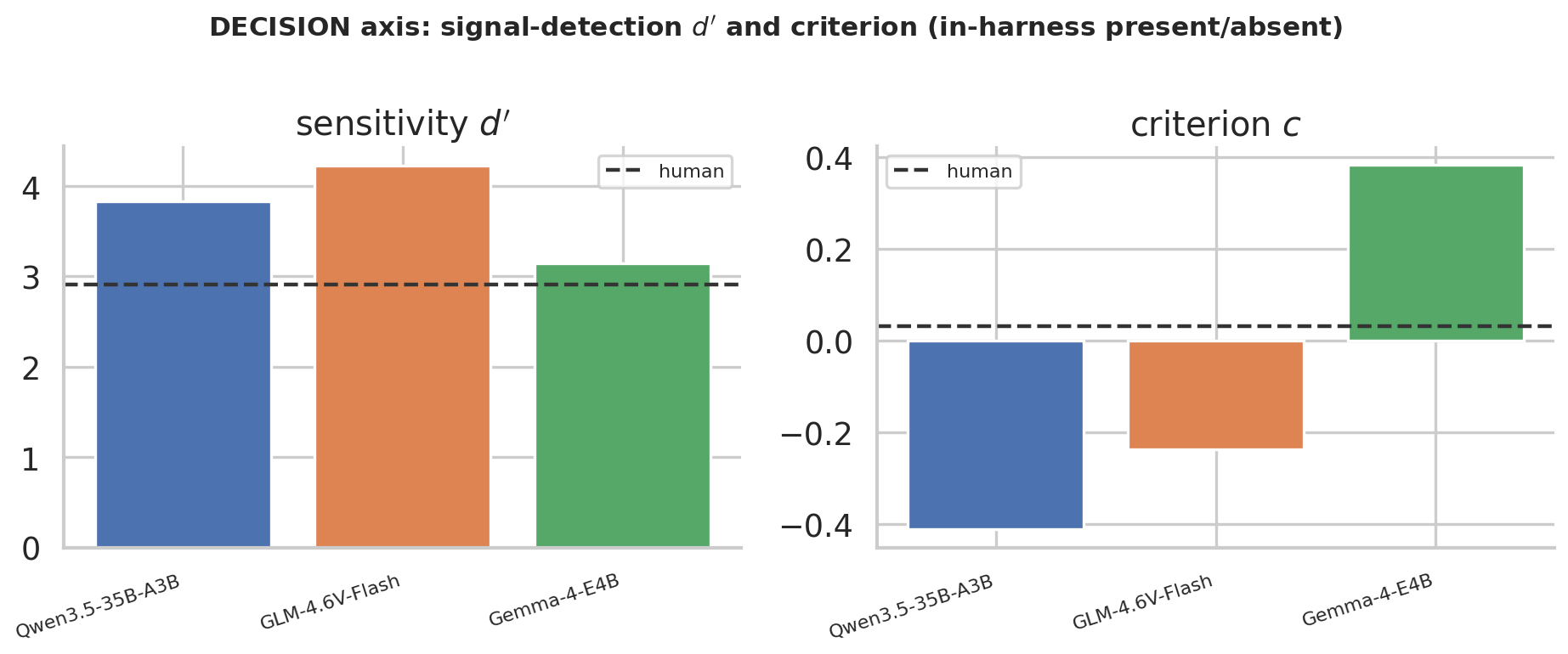}
\caption{Signal-detection sensitivity ($d'$) and criterion ($c$) for the in-harness present/absent
decision, per model against the human reference.}\label{fig:dec}\end{figure}
 
All three models detect targets near ceiling (target-present existence accuracy
\QwenTPexist/\GlmTPexist/\GemmaTPexist) with a comparable false-positive bias on absent scenes
(\QwenYesBias/\GlmYesBias/\GemmaYesBias), so subsequent search divergences cannot be attributed to
detection failure and the existence-passed sets are comparable across models. In the agentic task the
present/absent decision is highly sensitive for every model ($d'$
\QwenGPDprime/\GlmGPDprime/\GemmaGPDprime, against the human \HumanDprime), establishing that the
\emph{decision} axis is human-or-better. Model $d'$ is elicited from the agentic \textsc{found}/\textsc{absent} terminations and human $d'$ from the recorded gamepad response; this is a deliberate elicitation asymmetry rather than a confound, as both index the same present/absent judgment. Spatial targeting is measured in-harness (Sec.~\ref{sec:finding}), where the coordinate convention is fixed in the prompt; the resulting first-saccade rate is corroborated by high eventual success (TFP-end \QwenGPTFPend/\GlmGPTFPend/\GemmaGPTFPend) and low median fixation counts (\QwenGPNumFixTP/\GlmGPNumFixTP/\GemmaGPNumFixTP), and is stable across hit tolerances (Table~\ref{tab:tol}), so it is not an artefact of coordinate parsing.

\section{Finding axis}\label{sec:finding}
\begin{table}[ht]\centering
\caption{Outcome by foveation condition; each entry is Qwen / GLM / Gemma against the human reference
(first row). TFP@1 and TFP-end follow Eq.~\eqref{eq:tfp}; NumFix entries are per-group medians; the
last two columns give target-absent declared-absent rate and fixation-density CC with the human map.}
\label{tab:outcome-supp}
\setlength{\tabcolsep}{4pt}
\resizebox{\linewidth}{!}{\begin{tabular}{l *{18}{c}}
\toprule
& \multicolumn{3}{c}{TFP@1} & \multicolumn{3}{c}{TFP-end} & \multicolumn{3}{c}{NFix-TP}
& \multicolumn{3}{c}{NFix-TA} & \multicolumn{3}{c}{TA decl.-abs} & \multicolumn{3}{c}{dens.\ CC} \\
\cmidrule(lr){2-4}\cmidrule(lr){5-7}\cmidrule(lr){8-10}\cmidrule(lr){11-13}\cmidrule(lr){14-16}\cmidrule(lr){17-19}
condition & Q & G & Gm & Q & G & Gm & Q & G & Gm & Q & G & Gm & Q & G & Gm & Q & G & Gm \\
\midrule
\textbf{human} & \multicolumn{3}{c}{\textbf{0.49}} & \multicolumn{3}{c}{\textbf{0.93}}
& \multicolumn{3}{c}{\textbf{3}} & \multicolumn{3}{c}{\textbf{5}}
& \multicolumn{3}{c}{---} & \multicolumn{3}{c}{---} \\
\midrule
sharp                  & 0.97 & 0.97 & 0.80 & 0.97 & 0.98 & 0.93 & 2 & 2 & 3 & 1 & 4 & 6 & 0.94 & 0.91 & 0.96 & 0.55 & 0.54 & 0.49 \\
\textbf{GP} & 0.97 & 0.97 & 0.80 & 0.98 & 0.98 & 0.93 & 2 & 2 & 3 & 1 & 4 & 5 & 0.93 & 0.91 & 0.96 & 0.58 & 0.63 & 0.50 \\
k8                     & 0.93 & 0.84 & 0.64 & 0.95 & 0.89 & 0.86 & 3 & 2 & 3 & 2 & 4 & 6 & 0.84 & 0.77 & 0.81 & 0.65 & 0.45 & 0.26 \\
k16                    & 0.78 & 0.71 & 0.42 & 0.91 & 0.80 & 0.75 & 3 & 3 & 4 & 4 & 4 & 6 & 0.78 & 0.77 & 0.68 & 0.70 & 0.38 & 0.45 \\
k24                    & 0.56 & 0.49 & 0.28 & 0.81 & 0.65 & 0.65 & 3 & 3 & 6 & 6 & 4 & 6 & 0.72 & 0.68 & 0.58 & 0.74 & 0.46 & 0.29 \\
k32                    & 0.42 & 0.33 & 0.19 & 0.71 & 0.53 & 0.59 & 3 & 3 & 6 & 7 & 4.5 & 7 & 0.62 & 0.65 & 0.52 & 0.70 & 0.43 & 0.26 \\
k48                    & 0.28 & 0.18 & 0.10 & 0.56 & 0.36 & 0.47 & 5 & 4 & 6 & 7 & 4 & 7 & 0.46 & 0.65 & 0.54 & 0.49 & 0.33 & 0.25 \\
k128                   & 0.15 & 0.09 & 0.01 & 0.31 & 0.22 & 0.39 & 6 & 4 & 6 & 6 & 4 & 6 & 0.85 & 0.86 & 0.78 & 0.18 & 0.14 & 0.19 \\
crop                   & 0.04 & 0.04 & 0.00 & 0.20 & 0.11 & 0.16 & 3 & 3 & 5 & 2 & 3 & 5 & 0.68 & 0.50 & 0.67 & 0.25 & 0.11 & 0.15 \\
\bottomrule
\end{tabular}}
\end{table}
\begin{figure}[ht]\centering\includegraphics[width=0.68\linewidth]{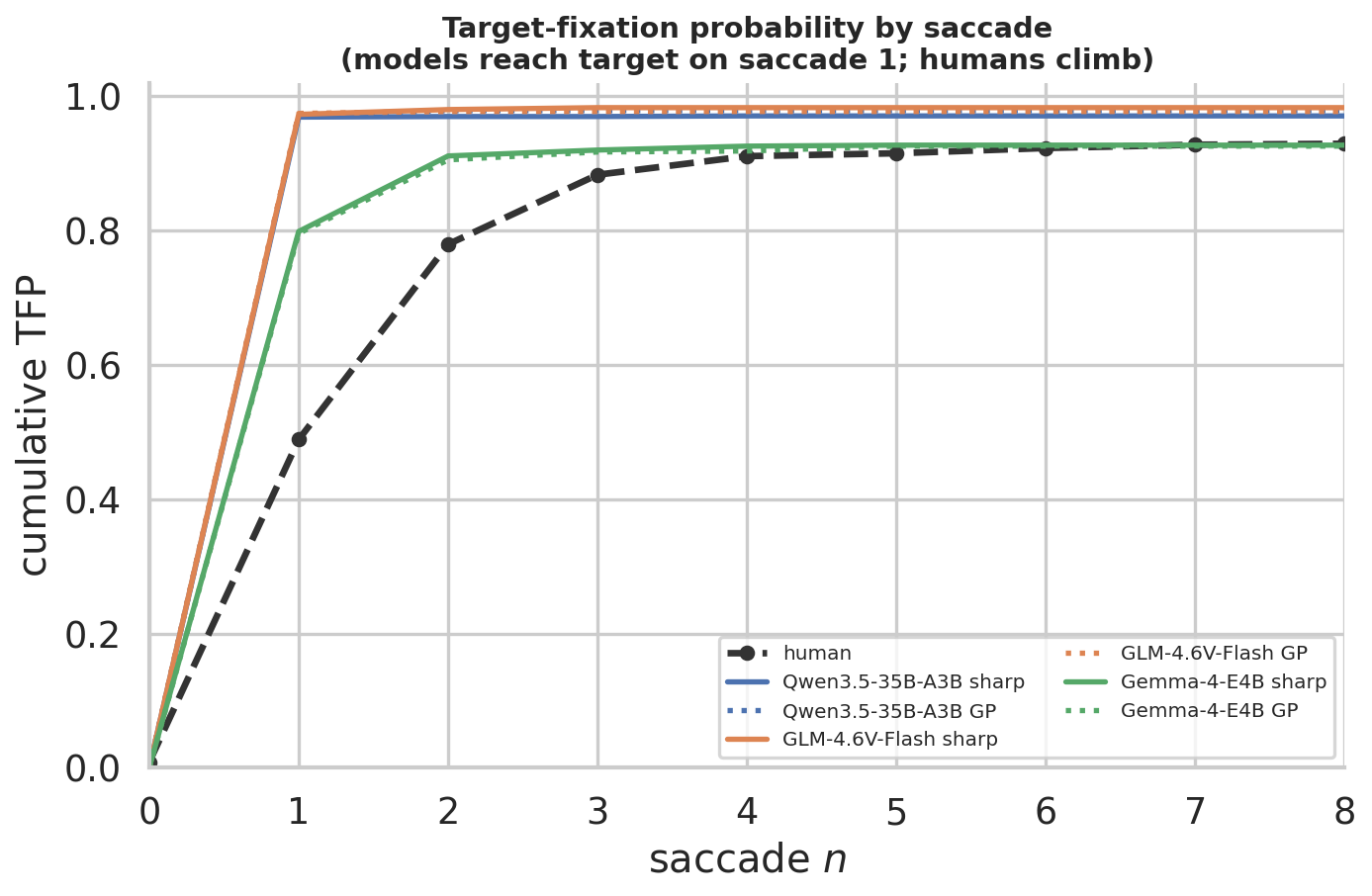}
\caption{Cumulative target-fixation probability by saccade (Eq.~\eqref{eq:tfp}) under \textsc{sharp}
(solid) and the human-matched GP (dotted) for the three models, against humans (dashed). The models
reach the target on the first saccade; human probability accrues over several fixations.}\label{fig:tfpc}\end{figure}
 
Under the human-matched condition every model fixates the target on the first saccade far more often than
humans (TFP@1 \QwenGPTFPone/\GlmGPTFPone/\GemmaGPTFPone\ versus \HumanTFPone) and reaches comparable
eventual success (TFP-end \QwenGPTFPend/\GlmGPTFPend/\GemmaGPTFPend\ versus \HumanTFPend), while issuing
no more fixations than humans (median NumFix \QwenGPNumFixTP/\GlmGPNumFixTP/\GemmaGPNumFixTP\ versus
\HumanNumFixTP). The advantage is therefore one of first-saccade efficiency rather than eventual
accuracy. Target-absent search extent varies across the three models: Qwen3.5-35B-A3B terminates after a single
fixation (median \QwenGPNumFixTA), GLM-4.6V-Flash searches longer (\GlmGPNumFixTA), and Gemma-4-E4B
searches to the human median (\GemmaGPNumFixTA\ versus \HumanNumFixTA). On the finding axis the models
are thus human-or-better throughout.

\section{Gaze axis}\label{sec:gaze-supp}
\begin{table}[ht]\centering
\caption{Intrinsic gaze signature: Cliff's $\delta$ against the human distribution (Eq.~\eqref{eq:cliffs};
$|\delta|>0.33$ in bold; sign is agent $-$ human) for every statistic and condition, for all three
models. Human medians: gaze entropy $1.58$ bits, saccade amplitude $8.57°$, refixation $0.00$, scanpath
length $19.4°$, center bias $10.0°$.}
\label{tab:signature}
\setlength{\tabcolsep}{4pt}\footnotesize
\resizebox{\linewidth}{!}{\begin{tabular}{llccccccc}
\toprule
model & condition & gaze entropy & saccade amp & refixation & scanpath len & center bias & turn angle & coverage \\
\midrule
Q & sharp & \textbf{$-$.68} & \textbf{$+$.52} & $+$.03 & $-$.27 & $-$.20 & $+$.03 & \textbf{$-$.75} \\
 & GP & \textbf{$-$.67} & \textbf{$+$.50} & $+$.04 & $-$.26 & $-$.18 & $+$.05 & \textbf{$-$.74} \\
 & k8 & \textbf{$-$.69} & $+$.20 & \textbf{$+$.36} & $-$.27 & $-$.05 & $+$.06 & \textbf{$-$.65} \\
 & k16 & \textbf{$-$.56} & $-$.03 & \textbf{$+$.49} & $-$.21 & $+$.01 & $+$.12 & \textbf{$-$.46} \\
 & k24 & \textbf{$-$.33} & $-$.08 & \textbf{$+$.52} & $-$.07 & $+$.02 & $+$.14 & $-$.18 \\
 & k32 & $-$.15 & $-$.07 & \textbf{$+$.50} & $+$.05 & $+$.00 & $+$.24 & $-$.03 \\
 & k48 & $+$.12 & $+$.01 & \textbf{$+$.54} & $+$.28 & $-$.04 & \textbf{$+$.45} & $+$.18 \\
 & k128 & $+$.33 & \textbf{$+$.73} & $+$.27 & \textbf{$+$.57} & $+$.17 & \textbf{$+$.67} & \textbf{$+$.46} \\
 & crop & $-$.24 & $+$.31 & $+$.26 & $-$.10 & \textbf{$-$.33} & \textbf{$+$.58} & $-$.11 \\
\midrule
G & sharp & \textbf{$-$.65} & \textbf{$+$.60} & $-$.12 & $-$.28 & $-$.29 & $-$.18 & \textbf{$-$.77} \\
 & GP & \textbf{$-$.65} & \textbf{$+$.61} & $-$.13 & $-$.28 & $-$.31 & $-$.21 & \textbf{$-$.77} \\
 & k8 & \textbf{$-$.65} & \textbf{$+$.43} & $+$.08 & $-$.28 & $-$.21 & $-$.06 & \textbf{$-$.71} \\
 & k16 & \textbf{$-$.53} & $+$.20 & $+$.30 & $-$.17 & $-$.05 & $+$.17 & \textbf{$-$.50} \\
 & k24 & \textbf{$-$.41} & $+$.02 & \textbf{$+$.45} & $-$.06 & $-$.02 & $+$.16 & \textbf{$-$.34} \\
 & k32 & $-$.26 & $-$.04 & \textbf{$+$.49} & $+$.07 & $-$.05 & $+$.19 & $-$.17 \\
 & k48 & $-$.10 & $-$.03 & \textbf{$+$.52} & $+$.25 & $+$.00 & $+$.19 & $+$.06 \\
 & k128 & $+$.01 & \textbf{$+$.42} & \textbf{$+$.43} & \textbf{$+$.61} & $+$.14 & \textbf{$+$.38} & $+$.29 \\
 & crop & \textbf{$-$.51} & $-$.08 & \textbf{$+$.65} & $-$.06 & \textbf{$-$.35} & $+$.31 & \textbf{$-$.37} \\
\midrule
Gm & sharp & $-$.28 & $+$.24 & $+$.13 & $-$.00 & $+$.06 & $+$.19 & $-$.28 \\
 & GP & $-$.27 & $+$.23 & $+$.13 & $+$.00 & $+$.06 & $+$.19 & $-$.26 \\
 & k8 & $-$.10 & $+$.01 & $+$.31 & $+$.14 & $+$.27 & $+$.24 & $+$.00 \\
 & k16 & $+$.19 & $+$.07 & $+$.27 & \textbf{$+$.35} & $+$.24 & $+$.30 & $+$.27 \\
 & k24 & \textbf{$+$.36} & $+$.16 & \textbf{$+$.38} & \textbf{$+$.56} & $+$.19 & \textbf{$+$.40} & \textbf{$+$.49} \\
 & k32 & \textbf{$+$.49} & $+$.20 & \textbf{$+$.41} & \textbf{$+$.66} & $+$.11 & \textbf{$+$.45} & \textbf{$+$.61} \\
 & k48 & \textbf{$+$.60} & $+$.30 & \textbf{$+$.45} & \textbf{$+$.76} & $-$.00 & \textbf{$+$.56} & \textbf{$+$.69} \\
 & k128 & \textbf{$+$.70} & \textbf{$+$.52} & \textbf{$+$.44} & \textbf{$+$.88} & $+$.02 & \textbf{$+$.62} & \textbf{$+$.82} \\
 & crop & $-$.08 & $+$.09 & \textbf{$+$.70} & \textbf{$+$.42} & $-$.19 & \textbf{$+$.48} & $+$.07 \\
\bottomrule
\end{tabular}
}
\end{table}
\begin{figure}[ht]\centering\includegraphics[width=\linewidth]{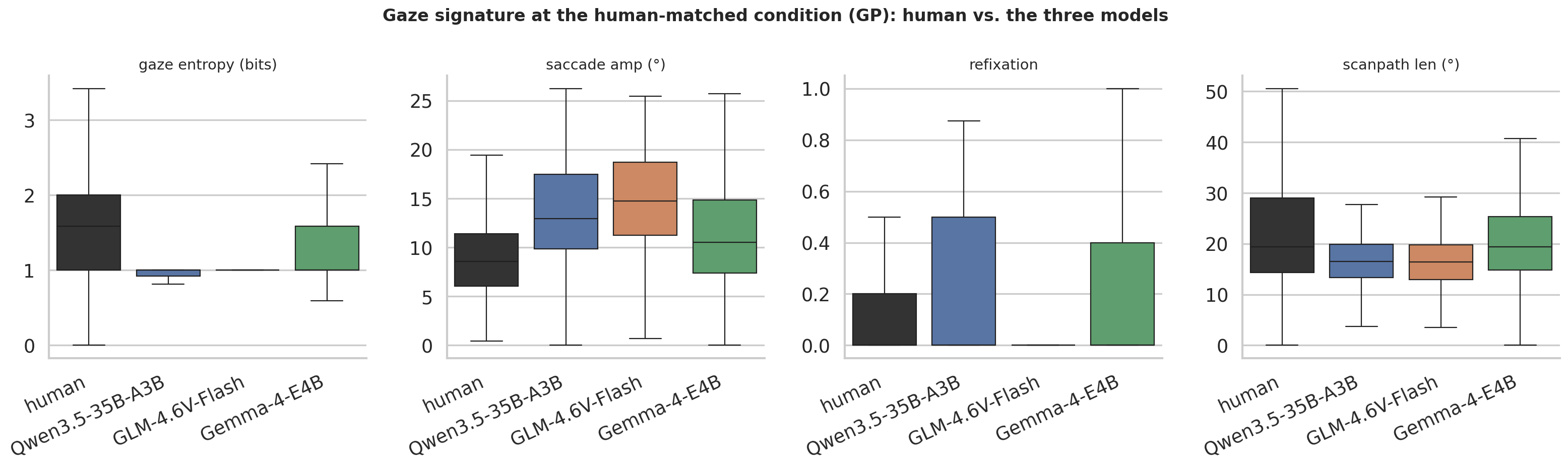}
\caption{Per-scanpath gaze-statistic distributions at the human-matched condition, human against the
three models. The reasoning-tuned models concentrate gaze (low entropy) and make large saccades; Gemma-4-E4B
lies closest to the human distribution on the entropy, amplitude and scanpath-length panels (Qwen3.5-35B-A3B is closest on refixation).}\label{fig:sig}\end{figure}
 
\begin{figure}[ht]
  \begin{subfigure}{\linewidth}\centering\includegraphics[width=0.82\linewidth]{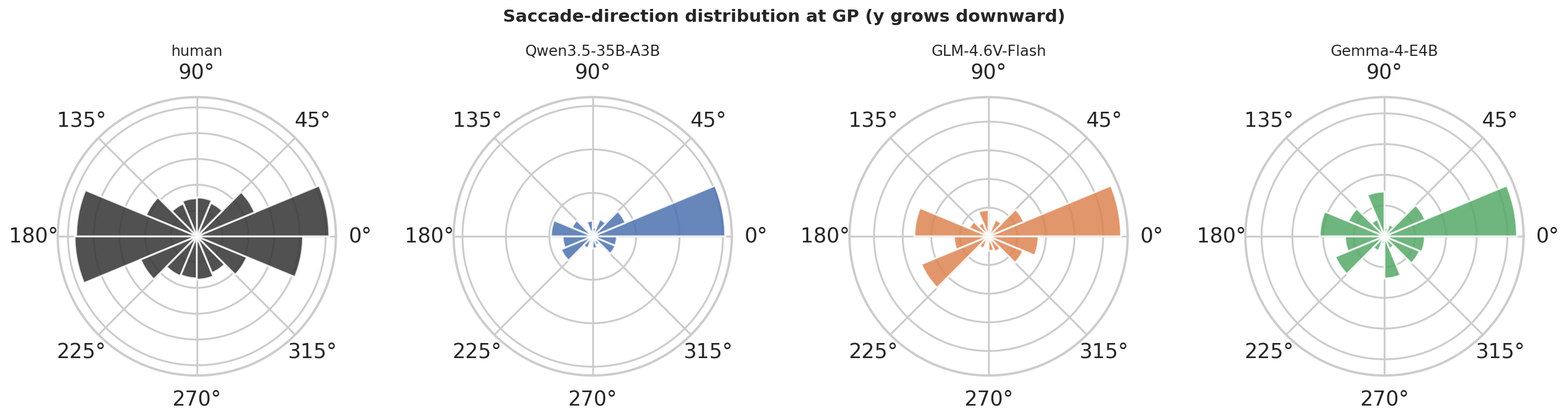}\caption{}\label{fig:polar}\end{subfigure}

  \begin{subfigure}{\linewidth}\centering\includegraphics[width=0.82\linewidth]{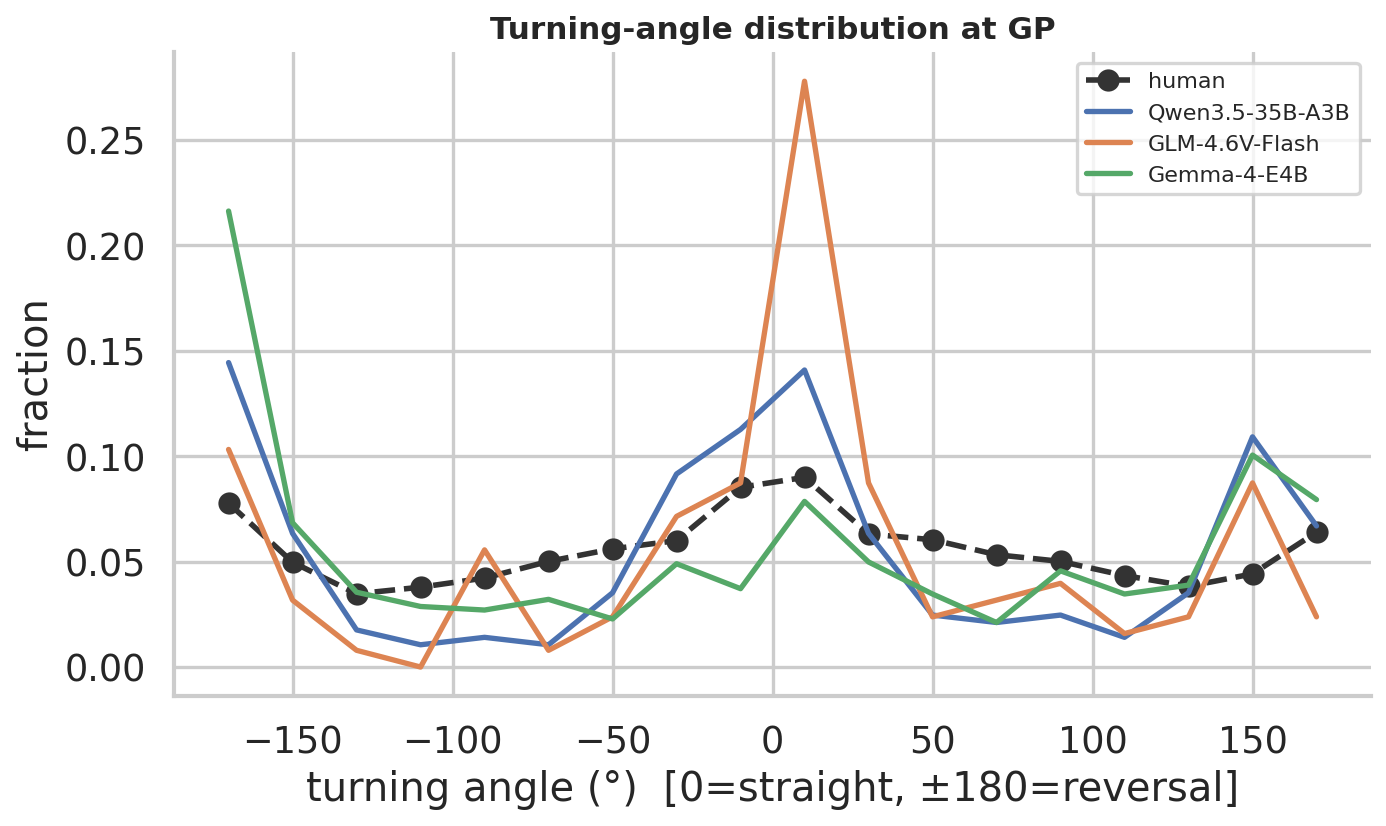}\caption{}\label{fig:turn}\end{subfigure}
  \caption{Distribution of (\subref{fig:polar}) saccade directions and (\subref{fig:turn}) turning angles
  at the human-matched condition. The models' directional and meander structure departs from the human
  reference, consistent with the entropy and amplitude effects.}\label{fig:dirturn}
\end{figure}

Where outcomes coincide with humans, the eye-movement process does not, and the deviation is in the same
direction for all three models: gaze entropy lies below the human value (Cliff's $\delta$
\QwenGPEntropyD/\GlmGPEntropyD/\GemmaGPEntropyD), indicating spatially concentrated sampling, and saccade
amplitudes exceed it (\QwenGPSaccD/\GlmGPSaccD/\GemmaGPSaccD), indicating direct jumps to the target. The
magnitude of the deviation varies across the three models, with Gemma-4-E4B's effects roughly half
those of the two reasoning-tuned models, but at this condition the sign is invariant, so the signature is shared rather than
idiosyncratic. The directional and turning-angle distributions (Fig.~\ref{fig:dirturn}) corroborate this
characterisation and show that it does not depend on the choice of summary statistic.
 
\begin{table}[ht]\centering
\caption{Scanpath similarity (ScanMatch, Sec.~\ref{sec:m-sim}) by condition: agent$\leftrightarrow$human
(AH) and agent$\leftrightarrow$agent (AA, cross-seed determinism) for each model. The
human$\leftrightarrow$human agreement ceiling is \HumanCeiling.}
\label{tab:similarity}
\setlength{\tabcolsep}{4pt}\small
\resizebox{\linewidth}{!}{\begin{tabular}{lcccccc}
\toprule
 & \multicolumn{2}{c}{Qwen3.5-35B-A3B} & \multicolumn{2}{c}{GLM-4.6V-Flash} & \multicolumn{2}{c}{Gemma-4-E4B} \\
condition & AH & AA & AH & AA & AH & AA \\
\midrule
\textbf{human ceiling} & \multicolumn{6}{c}{\textbf{0.53}} \\
sharp & 0.500 & 0.839 & 0.484 & 0.903 & 0.473 & 0.712 \\
GP & 0.505 & 0.843 & 0.485 & 0.913 & 0.469 & 0.713 \\
k8 & 0.527 & 0.814 & 0.491 & 0.868 & 0.462 & 0.697 \\
k16 & 0.521 & 0.776 & 0.476 & 0.800 & 0.405 & 0.594 \\
k24 & 0.479 & 0.696 & 0.445 & 0.712 & 0.324 & 0.477 \\
k32 & 0.420 & 0.597 & 0.402 & 0.613 & 0.275 & 0.424 \\
k48 & 0.332 & 0.499 & 0.320 & 0.495 & 0.240 & 0.373 \\
k128 & 0.244 & 0.407 & 0.266 & 0.423 & 0.217 & 0.397 \\
crop & 0.246 & 0.484 & 0.275 & 0.528 & 0.218 & 0.459 \\
\bottomrule
\end{tabular}
}
\end{table}
 
Cross-seed self-consistency exceeds the inter-observer ceiling for every model and follows the same
ordering as the intrinsic signature (agent$\leftrightarrow$agent ScanMatch
\QwenGPSelfConsist/\GlmGPSelfConsist/\GemmaGPSelfConsist\ versus the \HumanCeiling\ ceiling): each model
reproduces its own scanpath far more closely than two humans agree, Gemma-4-E4B least so.
Agent$\leftrightarrow$human similarity remains at or below the ceiling, so no model is more similar to a
human than two humans are to each other.

\section{Multivariate structure of the gaze signature}\label{sec:pca}
\begin{table}[ht]\centering
\caption{Principal components of the standardised five-statistic gaze signature across all groups, with
the loadings that render the axes interpretable.}
\label{tab:pca}\begin{tabular}{lcl}
\toprule
component & variance & dominant loadings \\
\midrule
PC1 & 45\% & $+0.62$ gaze entropy, $+0.62$ scanpath len, $+0.41$ center bias \\
PC2 & 30\% & $+0.69$ saccade amp, $-0.63$ refixation, $-0.35$ center bias \\
\bottomrule
\end{tabular}

\end{table}
 
A principal-component analysis of the per-group median signature vectors reduces the five statistics to
two interpretable axes (Table~\ref{tab:pca}); the leading component combines exploration extent
(scanpath length and gaze entropy) and the second contrasts saccade amplitude against refixation. In this
space every model, under every legible condition, occupies a region disjoint from the human reference
(main Fig.~5b), approaching it only under degradation severe enough that the target is no longer found.
That three architecturally distinct models from three families share this region is the multivariate
expression of the gaze divergence.

\section{Mechanism of the divergence}\label{sec:mech}
 
\begin{figure}[ht]\centering\includegraphics[width=0.56\linewidth]{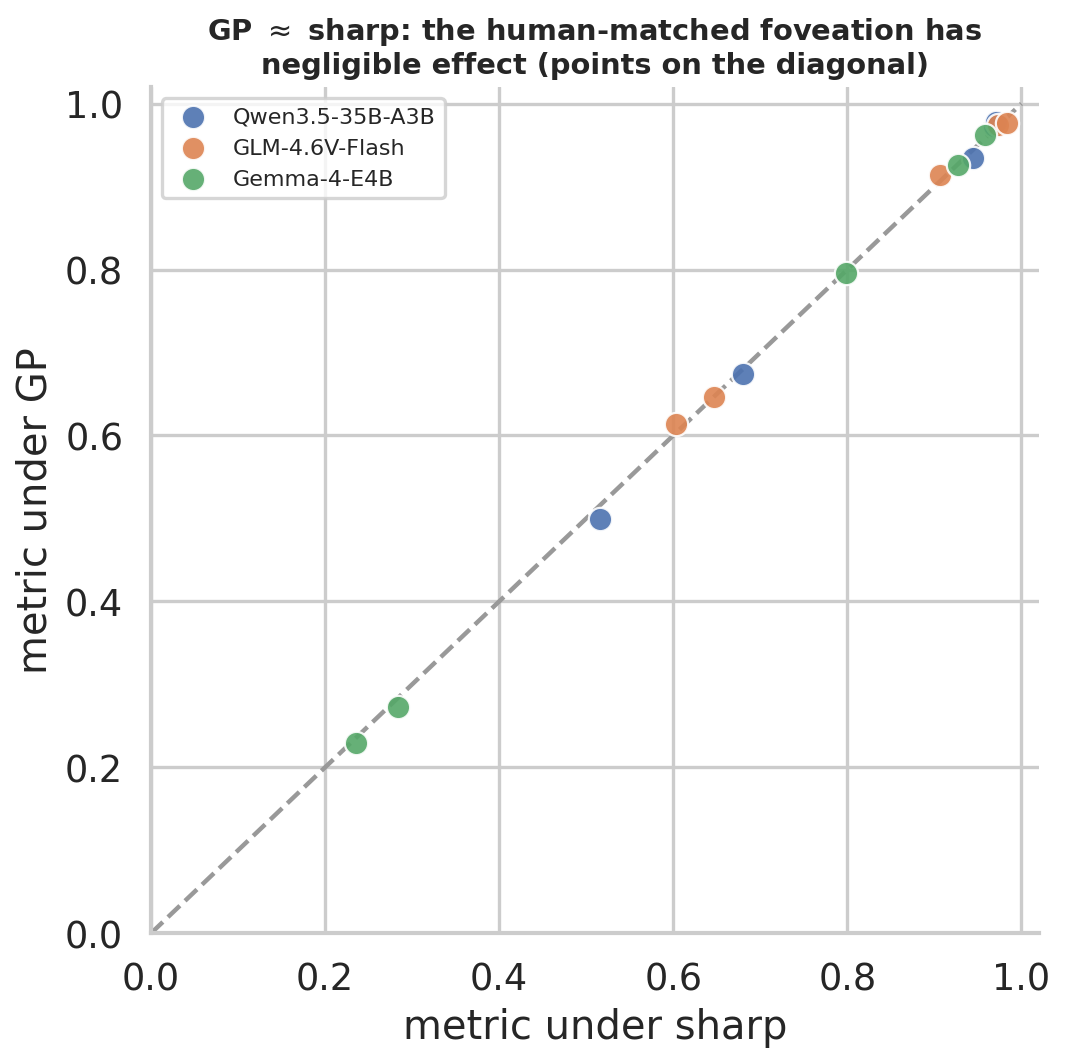}
\caption{Each statistic under \textsc{sharp} (abscissa) versus the human-matched GP condition (ordinate),
per model. Points lie on the identity line: the human-matched foveation changes behaviour negligibly.}\label{fig:gps}\end{figure}

\paragraph{The divergence is consistent with a single-pass architecture rather than an acuity effect.}
The human-matched condition is behaviourally indistinguishable from no foveation for every model and
statistic, even though it measurably degrades the periphery (Fig.~\ref{fig:gps}; e.g.\ first-saccade
targeting \QwenGPTFPone\ under GP versus \QwenSharpTFPone\ under \textsc{sharp}). A parallel vision encoder
resolves a legible frame in a single pass and saccades directly to the target; matching the retinal
\emph{input} therefore does not reconstruct the serial sampling constraint that produces human search.
 
\begin{figure}[ht]\centering\includegraphics[width=0.6\linewidth]{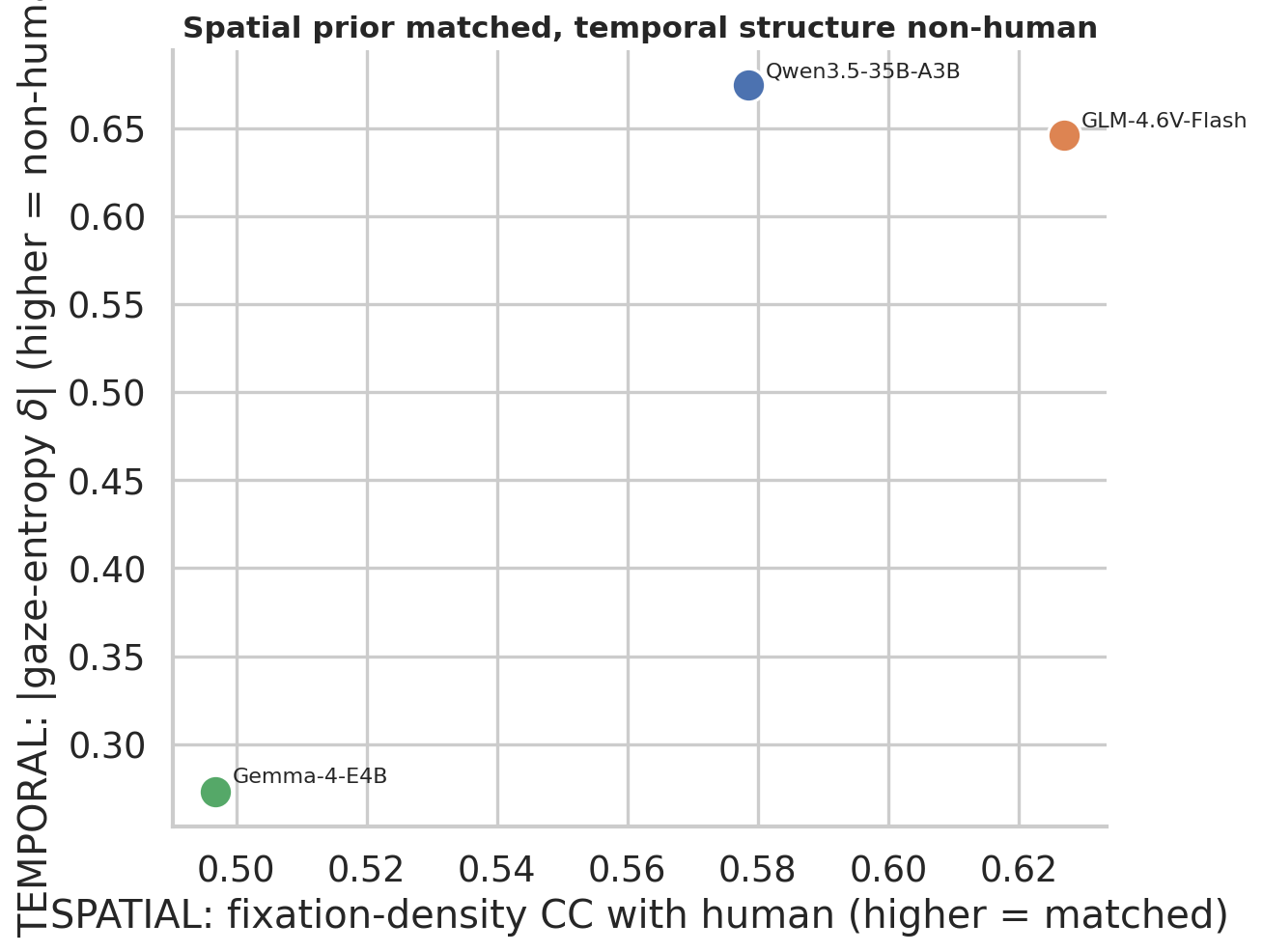}
\caption{Spatial agreement with humans (fixation-density CC, abscissa) against temporal divergence
(absolute gaze-entropy effect, ordinate). The models match \emph{where} humans look while diverging in
\emph{how} the looking unfolds.}\label{fig:spt}\end{figure}

\paragraph{The matched property is spatial; the divergent property is temporal.}
The models' spatial prior is approximately human: center bias is statistically indistinguishable from the
human value (sub-threshold $\delta$ in Table~\ref{tab:signature}) and fixation density correlates
positively with the human map (CC \QwenGPDensityCC/\GlmGPDensityCC/\GemmaGPDensityCC; full agreement measures in Table~\ref{tab:density}), so the models look
in broadly human-relevant places. What differs is the temporal organisation of looking: the order,
amplitude and determinism of fixations (Fig.~\ref{fig:spt}). The missing component is serial sampling,
not the spatial prior.

\begin{table}[ht]\centering
\caption{Fixation-density agreement with the human map (Sec.~\ref{sec:m-dens}) under \textsc{sharp} and the human-matched condition: linear correlation CC, normalised scanpath saliency NSS, and Kullback--Leibler divergence KL. CC and NSS (higher is closer) are highest for GLM-4.6V-Flash; KL (lower is closer) is lowest for Gemma-4-E4B. All three models correlate positively with the human map.}
\label{tab:density}
\begin{tabular}{lcccccc}
\toprule
 & \multicolumn{3}{c}{\textsc{sharp}} & \multicolumn{3}{c}{\textsc{geisler--perry}} \\
\cmidrule(lr){2-4}\cmidrule(lr){5-7}
model & CC & NSS & KL & CC & NSS & KL \\
\midrule
Qwen3.5-35B-A3B & 0.55 & 0.57 & 1.39 & 0.58 & 0.60 & 1.30 \\
GLM-4.6V-Flash  & 0.54 & 0.54 & 1.19 & 0.63 & 0.63 & 0.74 \\
Gemma-4-E4B     & 0.49 & 0.48 & 0.47 & 0.50 & 0.48 & 0.46 \\
\bottomrule
\end{tabular}

\end{table}
 
\begin{figure}[ht]\centering\includegraphics[width=0.66\linewidth]{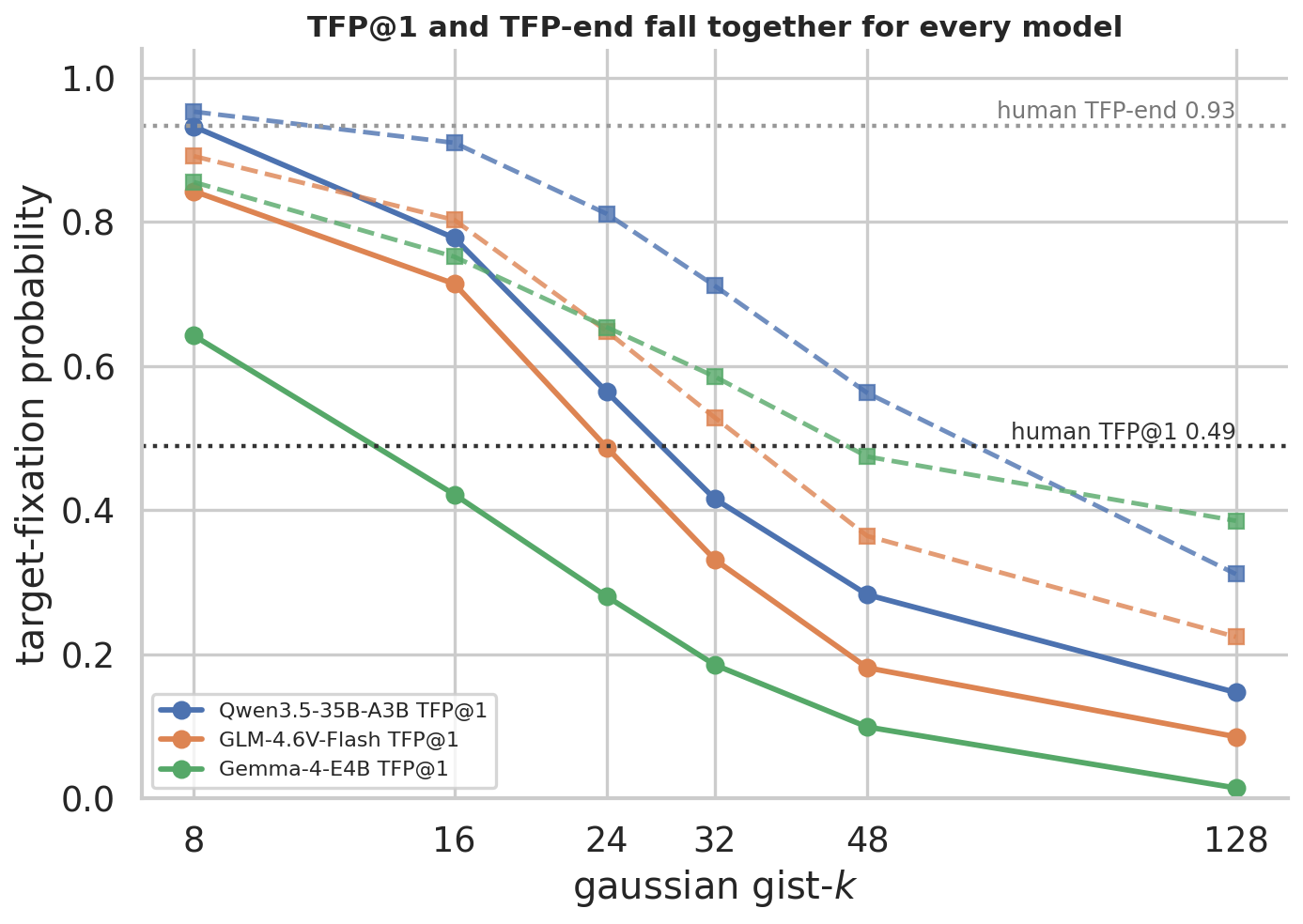}
\caption{First-saccade targeting (solid) and eventual success (dashed) as functions of the synthetic
degradation factor $k$, for the three models, with human reference levels. The two curves fall together;
no $k$ yields human-like search at human-like success.}\label{fig:gistS}\end{figure}

\paragraph{No degradation regime recovers human-like search.}
Sweeping the synthetic degradation does not produce a regime that is simultaneously human-like in
first-saccade targeting and in eventual success (Fig.~\ref{fig:gistS}): the two quantities decline
together. At the degradation level that lowers first-saccade targeting to the human rate
(\HumanTFPone), namely $k{=}32$ for Qwen3.5-35B-A3B and $k{=}16$ for Gemma-4-E4B, eventual success has
already fallen well below the human level (TFP-end \QwenKthirtytwoTFPend\ and \GemmaKsixteenTFPend\
respectively). There is no operating point at which the models search like humans and still find the
target.
 
\begin{figure}[ht]\centering\includegraphics[width=0.66\linewidth]{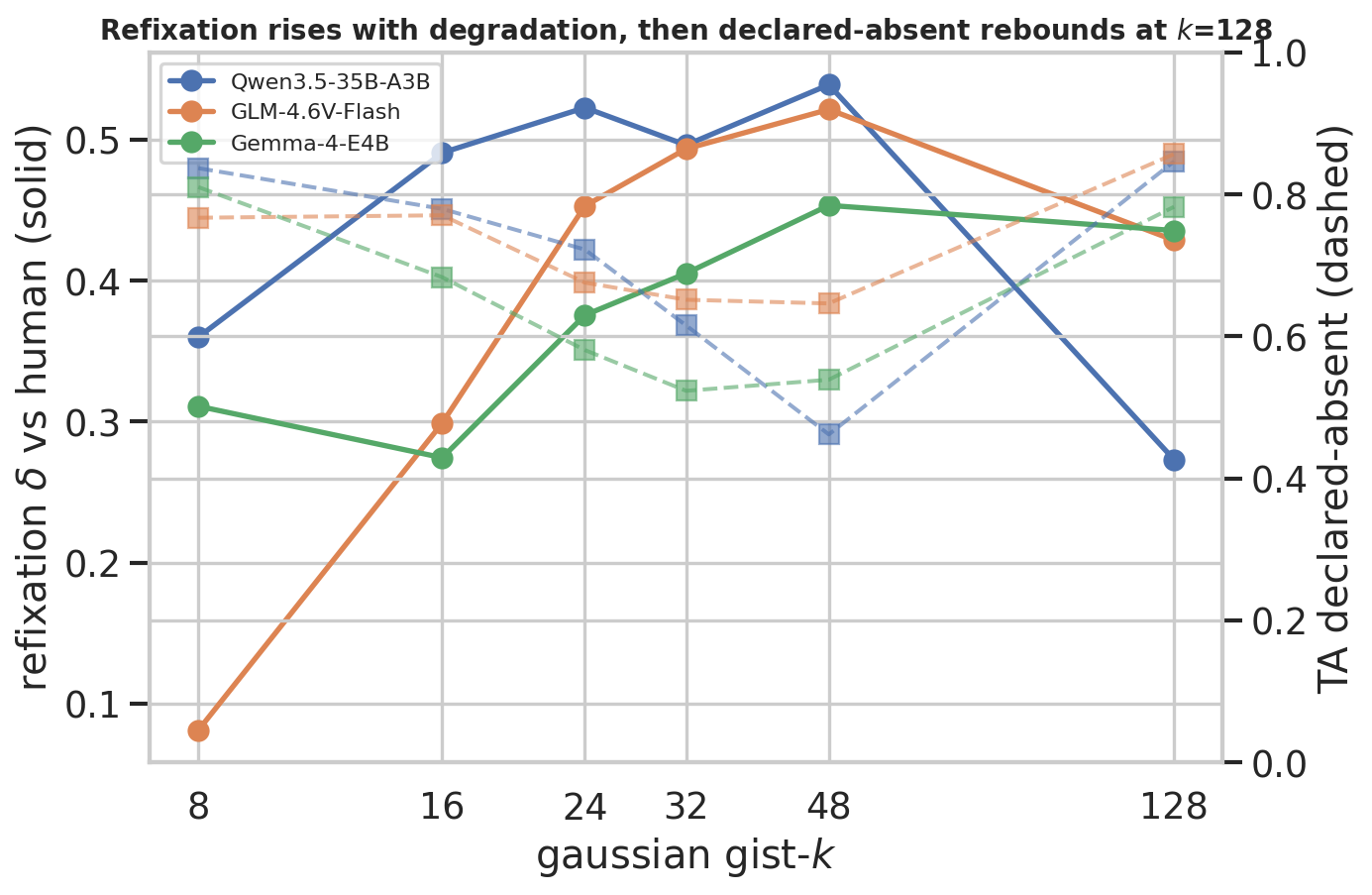}
\caption{As degradation increases, the refixation effect (solid) rises (the models revisit locations) and
then the declared-absent rate (dashed) rebounds at the most severe degradation as the models
default to an absent response.}\label{fig:thr}\end{figure}

\paragraph{Failure mode under vanishing evidence.}
As the periphery is degraded the models do not lengthen their search in a graceful, human-like manner.
Refixation first rises as cells are revisited, and at the most severe degradation the target-absent
declared-absent rate rebounds as the models default to an ``absent'' response
(Fig.~\ref{fig:thr}). Because one clause of the prompt encourages the use of glimpse memory, the absolute
refixation level is in part prompt-shaped; the failure signature is the trend across degradation, not its
absolute value. This pattern, rising refixation followed by default-absent termination, is model-internal and requires no human reference to detect.
 
\section{Robustness and cross-model variation}\label{sec:robust}
\begin{table}[ht]\centering
\caption{First-saccade targeting by eccentricity $\times$ target-size stratum (sharp), human against the
three models.}
\label{tab:strata}\begin{tabular}{lcc}
\toprule
stratum (n) & human TFP@1 & sharp TFP@1 (Q / G / Gm) \\
\midrule
near-small (44) & 0.54 & 0.96 / 0.97 / 0.76 \\
near-large (26) & 0.65 & 0.97 / 0.98 / 0.85 \\
far-small (26) & 0.37 & 1.00 / 0.98 / 0.64 \\
far-large (45) & 0.39 & 0.96 / 0.96 / 0.89 \\
\bottomrule
\end{tabular}

\end{table}
\begin{figure}[ht]\centering\includegraphics[width=0.6\linewidth]{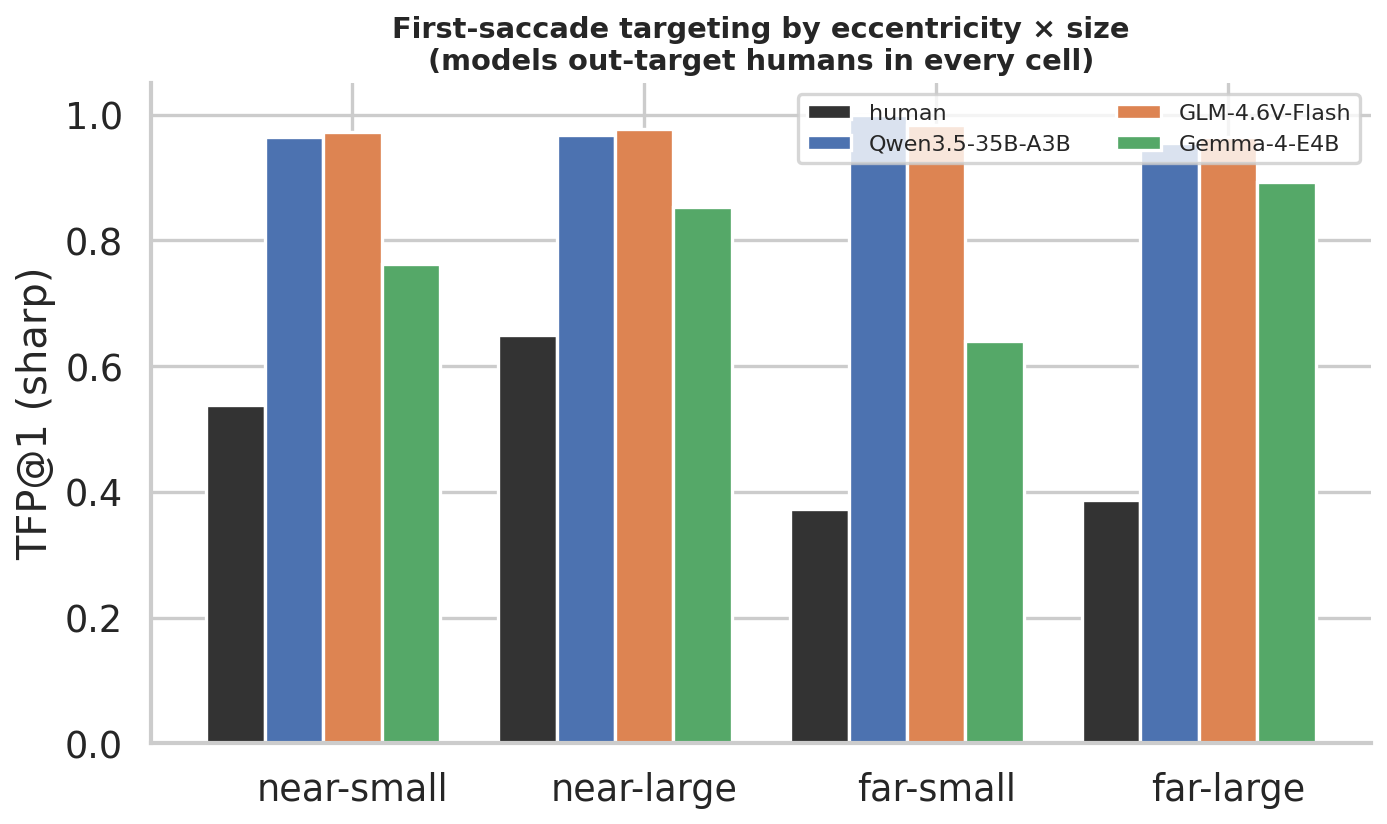}
\caption{First-saccade targeting per difficulty stratum; the models exceed humans in every eccentricity
$\times$ size cell, including the hardest.}\label{fig:strat}\end{figure}
 
The first-saccade advantage is not an artifact of easy targets: it holds in every eccentricity $\times$
size stratum, including the hardest (Table~\ref{tab:strata}). It is likewise insensitive to the
target-box tolerance: recomputing TFP@1 at $\pm0.5°$, $\pm1°$ and $\pm1.5°$ shifts any value by at most
the amounts in Table~\ref{tab:tol}, and the model--human gap holds at every tolerance.
 
\begin{table}[ht]\centering
\caption{Maximum absolute change in TFP@1 across box tolerances of $0.5$, $1$ and $1.5$ degrees, per
model.}
\label{tab:tol}\begin{tabular}{lc}
\toprule
model & max $|\Delta$TFP@1$|$ over $\pm0.5/1/1.5°$ \\
\midrule
Qwen3.5-35B-A3B & 0.063 \\
GLM-4.6V-Flash & 0.076 \\
Gemma-4-E4B & 0.095 \\
\bottomrule
\end{tabular}

\end{table}
 
\begin{table}[ht]\centering
\caption{Linear mixed-effects fixed effects against the human reference (Eq.~\eqref{eq:lmm}; $\Delta$
[95\% CI]) under \textsc{sharp} and the human-matched condition, with crossed scene and rater random
intercepts. Bold indicates a confidence interval that excludes zero.}
\label{tab:mixed}\begin{tabular}{llcc}
\toprule
model & metric & sharp $\Delta$ [95\% CI] & GP $\Delta$ [95\% CI] \\
\midrule
Qwen3.5-35B-A3B & gaze entropy & \textbf{$-0.62$ [$-0.69,-0.56$]} & \textbf{$-0.61$ [$-0.67,-0.54$]} \\
 & saccade amp. & \textbf{$+4.57$ [$+3.19,+5.95$]} & \textbf{$+4.53$ [$+3.15,+5.91$]} \\
 & NumFix & \textbf{$-1.53$ [$-1.94,-1.13$]} & \textbf{$-1.49$ [$-1.89,-1.08$]} \\
 & TFP@1 & \textbf{$+0.47$ [$+0.40,+0.54$]} & \textbf{$+0.48$ [$+0.41,+0.55$]} \\
\midrule
GLM-4.6V-Flash & gaze entropy & \textbf{$-0.60$ [$-0.67,-0.52$]} & \textbf{$-0.60$ [$-0.67,-0.52$]} \\
 & saccade amp. & \textbf{$+5.58$ [$+4.84,+6.32$]} & \textbf{$+5.62$ [$+4.88,+6.35$]} \\
 & NumFix & \textbf{$-1.51$ [$-1.89,-1.13$]} & \textbf{$-1.60$ [$-1.98,-1.22$]} \\
 & TFP@1 & \textbf{$+0.48$ [$+0.40,+0.56$]} & \textbf{$+0.48$ [$+0.40,+0.56$]} \\
\midrule
Gemma-4-E4B & gaze entropy & \textbf{$-0.23$ [$-0.33,-0.14$]} & \textbf{$-0.22$ [$-0.31,-0.12$]} \\
 & saccade amp. & \textbf{$+2.17$ [$+1.34,+3.00$]} & \textbf{$+2.20$ [$+1.37,+3.03$]} \\
 & NumFix & $-0.26$ [$-0.78,+0.27$] & $-0.03$ [$-0.55,+0.49$] \\
 & TFP@1 & \textbf{$+0.31$ [$+0.19,+0.42$]} & \textbf{$+0.30$ [$+0.19,+0.41$]} \\
\bottomrule
\end{tabular}

\end{table}
 
\begin{figure}[ht]\centering\includegraphics[width=0.6\linewidth]{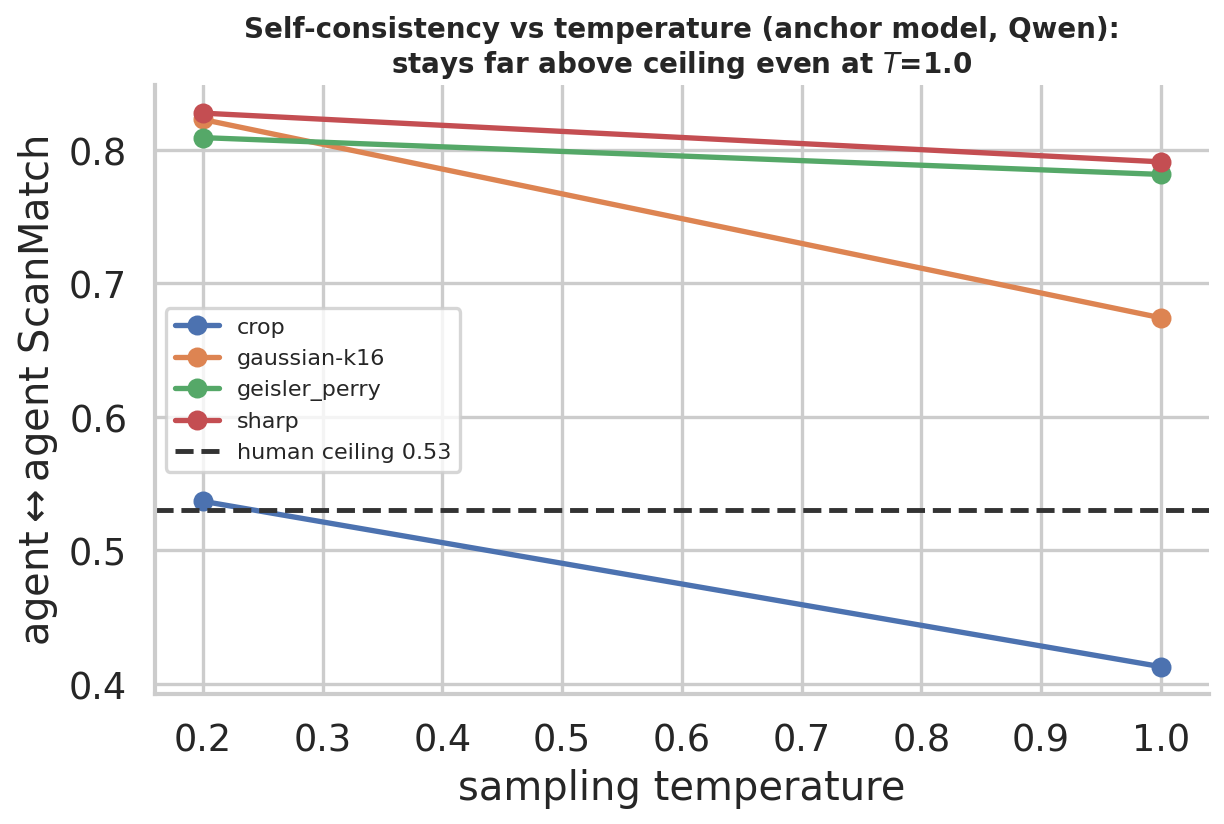}
\caption{Cross-seed self-consistency as a function of sampling temperature (anchor model, Qwen, core
conditions), against the human$\leftrightarrow$human ceiling \HumanCeiling.}\label{fig:temp}\end{figure}
 
The distributional effects survive the mixed-effects model of Eq.~\eqref{eq:lmm}, which controls for both
scene and rater (Table~\ref{tab:mixed}): the gaze-entropy, saccade-amplitude and first-saccade effects
have confidence intervals excluding zero for all three models. Two features of the cross-model ordering
are notable. First, Gemma-4-E4B's effects are the smallest on five of the seven gaze statistics (entropy, saccade
amplitude, scanpath length, center bias, coverage), making it the most human-like model overall; Qwen3.5-35B-A3B
is closest to humans on refixation and turning angle (Table~\ref{tab:signature}). Second, at the human-matched condition its search-extent effect is the single
fixed effect whose interval includes zero ($\Delta$\GemmaNumFixDelta), making its number of fixations
statistically indistinguishable from the human value. We interpret this ordering as a consequence of
weaker single-pass targeting (which forces additional, smaller, more variable fixations) rather than a
more human-like search strategy. The three models differ in architecture, training recipe (only
Gemma-4-E4B is not reasoning-tuned) and mixture-of-experts sparsity; these factors covary and cannot
be separated with three observations, so the ordering is reported descriptively. The high cross-seed
determinism is not an artifact of low-temperature decoding: in a temperature sweep on the anchor model (Qwen3.5-35B-A3B)
it remains well above the inter-observer ceiling even at temperature $1.0$ on the legible conditions
(Fig.~\ref{fig:temp}), and the deployed-temperature self-consistency of all three models
(Table~\ref{tab:similarity}) shows the same ordering.

\section{Data completeness}\label{sec:integrity}
All three models completed the full design (\NImages\ scenes, \NSeeds\ seeds and nine conditions), with
no trials excluded from the final analysis. Two protocol details bear on the integrity of the records. GLM-4.6V-Flash's one-shot detection responses were re-collected after an initial server interruption, recovering
its detection ceiling without affecting its search records. Gemma-4-E4B emitted its terminal
present/absent decision in a coordinate format requiring canonical normalisation before parsing; the
affected episodes were reconstructed from their logged turn sequences, each verified to reproduce the
recorded gaze path up to the decision, and the residual truncated episodes were re-collected under the
identical protocol. Neither procedure altered the other models' records, and all reported quantities are
computed from the recorded scanpaths under the single set of definitions given in
Sec.~\ref{sec:metrics}.

\section{Implications and scope}\label{sec:scope}
\paragraph{Evaluation metrics.} The dissociation has a direct methodological consequence. The models
match the human present/absent answer and partially match the human fixation-density map (CC
\QwenGPDensityCC/\GlmGPDensityCC/\GemmaGPDensityCC) while diverging on every temporal measure of the gaze
process. Because answer-alignment scores and single-shot saliency overlap are computed on outcomes or on
a time-collapsed spatial map, neither is sensitive to the axis on which the divergence occurs; a high
score on either is therefore necessary but not sufficient evidence of human-like vision, and certifying
process-level correspondence requires sequence-level, temporal measurement of the kind defined here.
 
\paragraph{Suitability as human-vision surrogates.} It follows that zero-shot multimodal models are
adequate surrogates for studies of \emph{outcome and spatial allocation} (detectability, approximate
region of interest, present/absent rates) and inadequate for studies of \emph{process and temporal
dynamics} (scanpath prediction, fixation counts and amplitudes, stopping behaviour, the time course of
evidence accumulation). Because the inadequacy is a shared property of the class rather than of any
instance, it is not removed by model selection within this family.
 
\paragraph{A null model for serial search.} Conversely, a system that attains human-or-better outcomes
\emph{without} a serial sampling bottleneck is a useful null model: contrasting a serial searcher (human
or model) against this parallel one-pass reference isolates the behaviours attributable to seriality
(eccentricity-dependent search cost, inspection-driven refixation, graded confirmatory stopping) from
those attributable to detection ability or spatial priors, which the null already reproduces.
 
\paragraph{Scope.} This study examined general-purpose models under a fixed foveation constraint. It did
not evaluate search-trained agentic models with learned zoom or tool-use policies, pointing-native or
frontier closed-source models, or an explicit probe of semantic guidance; nor did it run per-model
temperature sweeps beyond the anchor model. None of these bears on the three central findings, which
already hold across three distinct model families.

\clearpage
\section{Prompt}\label{sec:prompt}

\lstset{style=prompt}
\begin{lstlisting}
You are controlling a single eye that searches a photograph for a specific object.
You can only see clearly at the point you are currently looking; everything else is
blurred, with sharpness falling off the farther it is from your gaze, like human
peripheral vision. To inspect another region you must move your gaze there.

Each turn you receive the image as it currently looks from your gaze point. Earlier
turns show where you looked before and what you saw; use that history to decide where
to look next and to avoid re-checking the same spots.

Coordinates are normalized: x and y are each between 0.0 and 1.0. (0,0) is the
top-left corner, (1,1) the bottom-right; x grows rightward, y grows downward.

Your job is a PRESENT/ABSENT decision: is the target object in this image? On every turn
do exactly one of:
  - move your gaze to a new point to keep searching;
  - decide the target is PRESENT (FOUND): you can clearly see it at your gaze point;
  - decide the target is ABSENT: you are confident it is nowhere in the image.

Think briefly if you want, then end your reply with ONE directive line in EXACTLY one
of these forms and nothing after it:
  LOOK: x=<0..1>, y=<0..1>
  FOUND: x=<0..1>, y=<0..1>
  ABSENT
\end{lstlisting}

\end{document}